\documentclass{egpubl_arxiv}
\usepackage{eg2024}

\usepackage[T1]{fontenc}
\usepackage{dfadobe}

\usepackage{cite}
\BibtexOrBiblatex
\electronicVersion
\PrintedOrElectronic
\ifpdf \usepackage[pdftex]{graphicx} \pdfcompresslevel=9
\else \usepackage[dvips]{graphicx} \fi

\usepackage{egweblnk}

\usepackage{xcolor}
\usepackage{amssymb}
\usepackage{amsmath}
\usepackage{dsfont}
\usepackage{amsfonts}

\usepackage{pifont} 
\usepackage{booktabs}
\usepackage{multirow}
\usepackage{makecell}
\usepackage{adjustbox}
\usepackage{fontawesome} 
\usepackage{caption}
\usepackage{subcaption}
\usepackage{cleveref}
\usepackage{comment}
\usepackage{algorithm}
\usepackage{algorithmicx}
\usepackage{algpseudocode}

\newcommand{\etal       }     {{et~al.}}

\newcolumntype{H}{>{\setbox0=\hbox\bgroup}c<{\egroup}@{}}

\definecolor{darkred}{rgb}{0.7,0.1,0.1}
\definecolor{darkgreen}{rgb}{0.1,0.5,0.1}
\definecolor{cyan}{rgb}{0.7,0.0,0.7}
\definecolor{otherblue}{rgb}{0.1,0.4,0.8}
\definecolor{maroon}{rgb}{0.76,.13,.28}
\definecolor{burntorange}{rgb}{0.81,.33,0}
\definecolor{othergreen}{rgb}{0.29,0.49,0.07}
\definecolor{orange}{rgb}{1.0,0.65,0.0}

\definecolor{MZcolor}{rgb}{0.0,0.0,0.8}
\definecolor{CLcolor}{rgb}{0.0, 0.8, 0.75}
\definecolor{darkorange}{rgb}{1.0, 0.55, 0.0}
\definecolor{darkgreen}{rgb}{0.0, 0.55, 0.0}
\definecolor{darkpurple}{rgb}{0.0, 0.55, 0.8}
\definecolor{RMBcolor}{rgb}{0.25, 0.52, 0.96}
\definecolor{BMcolor}{rgb}{0.5, 0.1, 0.8}
\definecolor{VScolor}{rgb}{0.5, 0.8, 0.8}
\definecolor{PScolor}{rgb}{0.7, 0.7, 0.2}

\definecolor{cmarkcolor}{rgb}{0.49,0.74,0.49}
\definecolor{xmarkcolor}{rgb}{0.86,0.34,0.34}

\newcommand{\cmark}{\ding{51}}
\newcommand{\xmark}{\ding{55}}

\definecolor{opA}{rgb}{0.9,0.6,0.0}
\definecolor{opB}{rgb}{0.35,0.70,0.90}
\definecolor{opC}{rgb}{0.8,0.40,0.0}
\definecolor{opD}{rgb}{0.0,0.60,0.50} %
\definecolor{opE}{rgb}{0.8,0.6,0.7}
\definecolor{opF}{rgb}{0.,0.45,0.70}

\def \varcond {\qvec{c}}
\def \diffusevar {\qvec{x}}

\def\code#1{
    \ifx&#1&
        \xmark{}
    \else
        {\href{#1}{\faExternalLink}}
    \fi
}

\title[State of the Art on Diffusion Models for Visual Computing]{State of the Art on Diffusion Models for Visual Computing}

\begin{document}

\author[R. Po \& W. Yifan \& V. Golyanik \etal]
{
\parbox{\textwidth}
{\centering \vspace{-0.5cm}
R. Po$^{1\star}$~~W. Yifan$^{1\star}$~~V. Golyanik$^{2\star}$~~K. Aberman$^{3}$~~J. T. Barron$^{4}$~~A. Bermano$^{5}$~~E. Chan$^{1}$~~T. Dekel$^{6}$~~A. Holynski$^{4,7}$ \\
A. Kanazawa$^{7}$~~C. K. Liu$^{1}$~~L. Liu$^{8}$~~B. Mildenhall$^{4}$~~M. Nie{\ss}ner$^{9}$~~B. Ommer$^{10}$~~C. Theobalt$^{2}$~~P. Wonka$^{11}$~~G. Wetzstein$^{1}$
}
\\
\parbox{\textwidth}
{\centering
$^{1}$Stanford University~~$^{2}$MPI for Informatics and VIA Center~~$^{3}$Snap Inc.~~$^{4}$Google Research~~$^{5}$Tel Aviv University~~$^{6}$Weizmann Institute of Science
\\$^{7}$UC Berkley~~$^{8}$University of Pennsylvania~~$^{9}$TU Munich~~$^{10}$LMU Munich~~$^{11}$KAUST~~$^{\star}$Equal contribution
}
}

\teaser{
  \vspace{-1.2cm}
  \includegraphics[width=0.95\linewidth]{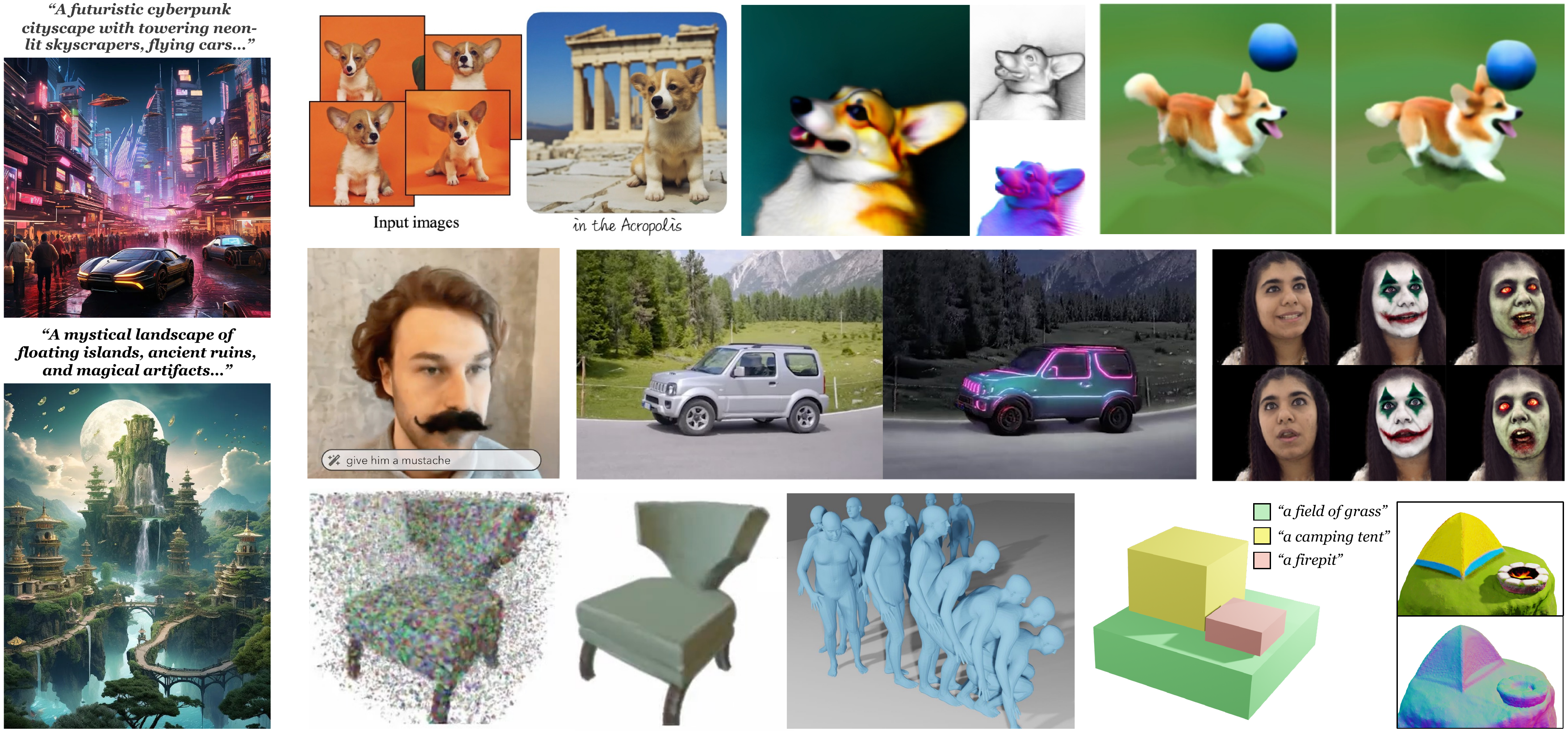}
  \centering\vspace{-0.3cm}
  \caption{
     This state-of-the-art report discusses the theory and practice of diffusion models for visual computing. These models have recently become the de-facto standard for image, video, 3D, and 4D generation and editing.
     Images adapted from~\cite{poole2022dreamfusion, dabral2022mofusion, Singer2023TextTo4DDS, muller2023diffrf, bar2022text2live, haque2023instruct, DeepFloyd2023, Po2023Compositional3S, Ruiz2022DreamBoothFT, Mendiratta2023, mj-web}
    \textcopyright 2023 IEEE.
    %
  }
  
  \label{fig:teaser}
}
\maketitle
\begin{abstract}
%
The field of visual computing is rapidly advancing due to the emergence of generative artificial intelligence (AI), which unlocks unprecedented capabilities for the generation, editing, and reconstruction of images, videos, and 3D scenes. In these domains, diffusion models are the generative AI architecture of choice. Within the last year alone, the literature on diffusion-based tools and applications has seen exponential growth and relevant papers are published across the computer graphics, computer vision, and AI communities with new works appearing daily on arXiv. This rapid growth of the field makes it difficult to keep up with all recent developments. The goal of this state-of-the-art report (STAR) is to introduce the basic mathematical concepts of diffusion models, implementation details and design choices of the popular Stable Diffusion model, as well as overview important aspects of these generative AI tools, including personalization, conditioning, inversion, among others. Moreover, we give a comprehensive overview of the rapidly growing literature on diffusion-based generation and editing, categorized by the type of generated medium, including 2D images, videos, 3D objects, locomotion, and 4D scenes. Finally, we discuss available datasets, metrics, open challenges, and social implications. This STAR provides an intuitive starting point to explore this exciting topic for researchers, artists, and practitioners alike.

\end{abstract}

\section{Introduction}
\label{sec:intro}

For decades, the computer graphics and 3D computer vision communities have been striving to develop physically accurate models to synthesize computer-generated imagery or infer physical properties of a scene from photographs. This methodology, which includes rendering, simulation, geometry processing, and photogrammetry, forms a cornerstone of several industries including visual effects, gaming, image and video processing, computer-aided design, virtual and augmented reality, data visualization, robotics, autonomous vehicles, remote sensing, among others.

The emergence of generative artificial intelligence (AI) marks a paradigm shift for visual computing. Generative AI tools enable the generation and editing of
photorealistic and stylized images, videos, or 3D objects with little more than a text prompt or high-level user guidance as input. These tools automate many laborious processes in visual computing that had previously been reserved for experts with specialized domain knowledge, making them more broadly accessible.

The unprecedented capabilities of generative AI have been unlocked by foundation models for visual computing, such as Stable Diffusion~\cite{rombach2022high}, Imagen~\cite{saharia2022photorealistic}, Midjourney~\cite{midjourney}, or DALL-E 2\cite{openaiDALLE} and DALL-E 3 \cite{openaiDALLE3}. Trained on hundreds of millions to billions of text--image pairs, these models have ``seen it all'' and, with an estimated few billion learnable parameters, are extremely large. After being trained on a massive cloud of high-end graphics processing units (GPUs), these models form the foundation of the aforementioned generative AI tools. The networks commonly used for image, video, and 3D object generation are typically variants of convolutional neural network (CNN)--based diffusion models that are combined in a multi-modal manner with text computed via transformer-based architectures, such as CLIP~\cite{radford2021learning}.

While much of the successful development and training of foundation models for 2D image generation has come from well-funded industry players using a massive amount of resources, there is still room for the academic community to contribute in major ways to the development of these tools for graphics and vision. For instance, it is not clear how to extend existing image foundation models to other, higher-dimensional domains, like video and 3D scene generation. This is largely due to the lack of certain types of training data. The web, for example, contains billions of 2D images but much fewer instances of high-quality and diverse 3D objects or scenes. Moreover, it is not obvious how to scale 2D image generation architectures to handle higher dimensions, as required for video, 3D scene, or 4D multi-view-consistent scene generation. Another example of an existing limitation is \emph{computation}: diffusion models are rather slow at inference time due to the large size of their networks and their iterative nature, and even though massive amounts of (unlabeled) video data exists on the web, current network architectures are often too inefficient to be trained in a reasonable amount of time or on a reasonable amount of compute resources.

Despite the remaining open challenges, recent developments have spurred an explosion of diffusion models for visual computing over the last year (see representative examples in Fig.~\ref{fig:teaser}). The goals of this state-of-the-art report (STAR) are to introduce the fundamentals of diffusion models, to present a structured overview of the many recent works focusing on applications of diffusion models in visual computing, and to outline open challenges.

This STAR is structured as follows: Sec.~\ref{sec:scope} outlines the scope and refers interested readers to surveys on closely related topics that are not covered here; Sec.~\ref{sec:fundamentals} gives an overview of the mathematical foundations of 2D diffusion; Sec.~\ref{sec:challenges} discusses the challenge of moving beyond 2D images towards video, 3D, and higher-dimensional diffusion models; Sec.~\ref{sec:video} outlines approaches to diffusion-based video synthesis and editing; Sec.~\ref{sec:spatial} summarizes recent approaches to 3D object and scene generation; Sec.~\ref{sec:4D_generation} includes a discussion on 4D spatio-temporal diffusion for multi-view consistent video, human motion and scene generation (e.g., using parametric human body models); Sec.~\ref{sec:data} includes a brief discussion on available training data; Sec.~\ref{sec:metrics} reviews metrics used for various generated content; Sec.~\ref{sec:open_challenges} outlines open challenges; Sec.~\ref{sec:social} discusses societal implications and ethical concerns; and Sec.~\ref{sec:conclusion} concludes the STAR.

\section{Scope of this STAR}\label{sec:scope}
In this STAR, we focus on recent advances in applications of diffusion models in visual computing. Specifically, we discuss the role of diffusion models in the context of generating and editing images, videos, 3D objects or scenes, and multi-view consistent 4D dynamic scenes. 
We start by laying down the mathematical underpinnings of diffusion models. This includes a brief introduction to the general diffusion process as applied to 2D images.
We then delve into how these techniques enable generative modeling of high-dimensional signals, as we provide a comprehensive overview of the leading methodologies of diffusion models for video, 3D and 4D data.
This report aims to highlight techniques leveraging diffusion models to tackle problems on data beyond the image domain. With that in mind, we do not cover each and every method only applicable to 2D data. We also do not discuss works that leverage generative pipelines other than diffusion models.

\paragraph*{Related Surveys.}
Other generative methods, such as GANs, are closely related to diffusion models. However, we consider them to be beyond the scope of this report. We refer readers to~\cite{gui2021review} for an in-depth discussion on GANs and~\cite{bond2021deep,li2023generative,ShiPengXu2022} for a broader review of other generative methods, or the use of different generative model architectures for multi-modal image synthesis and editing~\cite{zhan2023mise}. Recently, the term \textit{foundation model} has become analogous to a diffusion model trained on internet-scale data image~\cite{rombach2022high}. While this report discusses methods that leverage such models, please refer to~\cite{bommasani2021opportunities} for an introduction and overview of foundation models in the context of natural language processing, visual computing, and other domains. Last but not least, the explosive advances in text-to-image (T2I) generation has led to an intrinsic link between large language models (LLMs) and diffusion models; interested readers can refer to~\cite{Zhao2023ASO} for a comprehensive survey on LLMs.


\paragraph*{Selection Scheme.} This report covers papers published in the proceedings of major computer vision, machine learning, and computer graphics conferences, as well as preprints released on arXiv (2021--2023). The authors of this report have selected papers based on their relevance to the scope of this survey, as we aim to provide a comprehensive overview of the rapid advances in diffusion models in the context of visual computing. However, though this report serves as a list of state-of-the-art methods in a specific domain, we do not claim completeness and highly recommend that readers refer to cited works for in-depth discussions and details.


\begin{figure}[t]
    \centering
    \includegraphics[width=\linewidth]{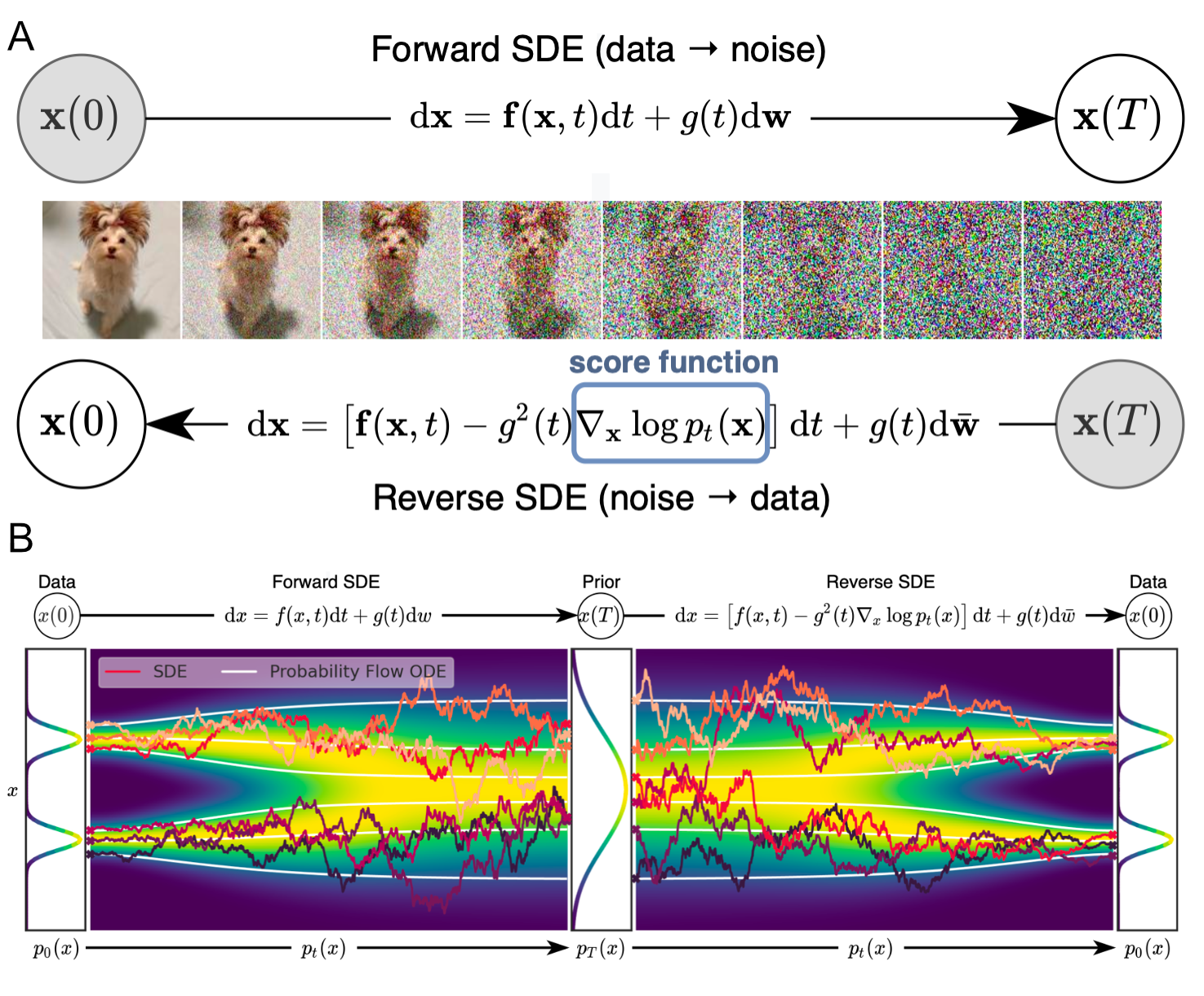}
    \caption{\textbf{Diffusion Process.} (A) The forward SDE transforms images to noise. The forward SDE can be reversed~\cite{anderson1982reverse} if we can predict the score function, enabling image synthesis. (B) The distributions of images and noise are linked with stochastic trajectories, modeled by SDEs, and deterministic trajectories, modeled by a probability flow ODE. Figures adapted from~\cite{song2020score}.}
    \label{fig:fundamentals_sdes}
\end{figure}

\section{Fundamentals of Diffusion Models}
\label{sec:fundamentals}

In this section, we give a concise overview of the fundamentals of diffusion models. We introduce the mathematical preliminaries, discuss a practical implementation using the popular Stable Diffusion model as an example, and then overview important concepts for conditioning and guidance, before discussing concepts related to inversion, image editing, and customization. This section covers a large amount of references, so we focus on giving the reader a clear and high-level overview of the most important concepts of diffusion-based generation and editing in the context of 2D images.

\subsection{Mathematical Preliminaries}


\newcommand{\qvec}[1]{\textbf{\textit{#1}}} 

Assume we are given a training dataset of examples where each example in the data is drawn independently from an underlying data distribution $p_{\text{data}}(\diffusevar)$. We desire to fit a model to $p_{\text{data}}(\diffusevar)$ so that we can synthesize novel examples by sampling from this distribution.

The general idea behind inference with denoising diffusion models is to sequentially denoise samples of random noise into samples from the data distribution. Consider a range of noise levels, $\sigma_{\text{max}} > \ldots > \sigma_{0} = 0$, and the corresponding noisy image distributions $p(\diffusevar, \sigma)$ defined as the distribution of adding Gaussian i.i.d. noise with variance $\sigma^2$ to the data. For sufficiently large $\sigma_{\text{max}}$, the noise almost completely obscures the data and $p(\diffusevar, \sigma_{\text{max}})$ is practically indistinguishable from Gaussian noise. Thus, we can sample an initial noise image $\diffusevar_T \sim \mathcal{N}(0, \sigma_{\text{max}}^2)$ and sequentially denoise it such that at every step, $\diffusevar_i \sim p(\diffusevar, \sigma_i)$. The endpoint of this sampling chain, $\diffusevar_0$ is distributed according to the data.

However, instead of thinking about the denoising through a discrete collection of noise levels, it is useful to think of the noise level as a continuous, time-dependent function $\sigma(t)$ (a common choice is $\sigma(t) = t$). The noisy image sample $\diffusevar$ can move continuously through noise levels, following a trajectory—either forward in time, gradually adding noise, or backwards in time, gradually removing noise (see Fig.~\ref{fig:fundamentals_sdes} (B)).

Song et al.~\cite{song2020score} introduce a stochastic differential equation (SDE) framework to model these trajectories. Ordinary differential equations give us tools to solve initial value problems -- given an initial state and a differential equation that describes a function, we can solve for the function at a different time. As an example, given an object’s initial position and known velocity, we can solve for the object’s position at any time in the future. As Fig.~\ref{fig:fundamentals_sdes} (A) depicts, in much the same way, noising an image can be thought of as picking an initial image $\diffusevar_0$ from the image domain and solving a differential equation forwards in time; denoising an image can be thought of as picking an initial noise image $\diffusevar_T \sim \mathcal{N}(0, \sigma_{\text{max}}^2)$ and solving a differential equation backwards in time.



The gradual corruption of an image with noise over time is a diffusion process that can be modeled by an It{\^o}
stochastic differential equation (SDE; Eq.~\ref{eq:ito_sde})  \cite{Itou1950, Itou1951}, where $\mathbf{f}(\cdot,t) : \mathbb{R}^d \rightarrow \mathbb{R}^d$ is a vector-valued function known as the drift coefficient, $g(\cdot) : \mathbb{R} \rightarrow \mathbb{R}$ is a scalar-valued function known as the diffusion coefficient, and $\qvec{w}$ is the standard Wiener process:
\begin{equation}
    \text{d}\diffusevar = \mathbf{f}(\diffusevar, t)\text{d}t + g(t)\text{d}\qvec{w}.
    \label{eq:ito_sde}
\end{equation}
Implementing a diffusion model requires selecting $\mathbf{f}$ and $g$, and several specific choices have been explored by \cite{song2020score}. The choices of $\mathbf{f}(\diffusevar, t) = 0$ and $g(t) = \sqrt{2 \sigma(t) \frac{\text{d}\sigma(t)}{\text{d}t}}$ yield an SDE that describes noising an image by adding Gaussian noise of variance $\sigma^2 (t)$. This SDE is known as the Variance Exploding SDE (Eq.~\ref{eq:ve_forward}), so-called because the variance continuously increases with increasing $t$. Noising an image can be thought of as selecting an initial clean image and solving Eq.~\ref{eq:ve_forward} forward in time as
\begin{equation}
    \text{d}\diffusevar = \sqrt{2 \sigma(t) \frac{\text{d}\sigma(t)}{\text{d}t}} \text{d} \qvec{w}.
    \label{eq:ve_forward}
\end{equation}

The Variance Exploding SDE has the closed-form solution
\begin{equation}
    p(\diffusevar_t |\diffusevar_0) = \mathcal{N}\left( \diffusevar_t; \diffusevar_0, \left[ \sigma^2 (t) - \sigma^2 (0) \right] \mathbf{I} \right). \label{eq:ve_closed_form}
\end{equation}
In other words, to obtain a noisy image at timestep $t$, all we need to do is add Gaussian noise of variance $\sigma^2 (t) - \sigma^2 (0)$.

The work of Anderson~\cite{anderson1982reverse} enables the discovery of an SDE that reverses a diffusion process. Applied to Eq.~\ref{eq:ve_forward}, this produces a reverse-time SDE
\begin{equation}
    \text{d}\diffusevar = - 2 \sigma(t) \frac{\text{d}\sigma (t)}{\text{d}t} \nabla_\diffusevar \log p(\diffusevar; \sigma(t)) \text{d} t + \sqrt{2 \sigma(t) \frac{\text{d} \sigma (t)}{\text{d} t}} \text{d} \qvec{w}. \label{eq:ve_reverse}
\end{equation}

In Eq.~\ref{eq:ve_reverse}, $\nabla_\diffusevar \log p(\diffusevar; \sigma(t))$ is known as the \textit{score function}, a vector field that points towards regions of higher data likelihood. To solve Eq.~\ref{eq:ve_reverse}, we need to predict the score function with a neural network. Remarkably, for a denoiser function $D$ that minimizes the $L_2$ denoising error $\mathbb{E}_{\qvec{y}\sim p_\text{data}} \mathbb{E}_{\qvec{n} \sim \mathcal{N}(\mathbf{0}, \sigma^2 \mathbf{I})}\|D(\qvec{y}+\qvec{n}; \sigma) - \qvec{y}\|^2_2$, the score function can be easily obtained from the model output as $\nabla_\diffusevar \log p(\diffusevar; \sigma(t)) = (D(\diffusevar;\sigma) - \diffusevar)/\sigma^2$. This means that by simply training a model to denoise images, we can extract a prediction of the score function. This is known as Denoising Score Matching and the above relationship was derived by~\cite{vincent2011connection}. Note that it is common to parameterize the neural network with parameters $\phi$ or $\theta$, as a noise-prediction network $\epsilon_\phi$, rather than as a denoiser $D_\theta$; however, each is easily recoverable from the other as $\epsilon_\phi(\diffusevar; \sigma) = \diffusevar - D_\theta(\diffusevar; \sigma)$.

In order to sample an image, we need only start with some initial $\diffusevar_T \sim \mathcal{N}(0, \sigma_{\text{max}}^2 \mathbf{I})$ and we can solve Eq.~\ref{eq:ve_reverse} backwards in time to arrive at a sample from the $p_\text{data}(\diffusevar)$. However, as is the case with ordinary differential equations (ODE), only a small subset has closed-form solutions. Fortunately, as an alternative, we can approximate the solution to the SDE numerically.

Euler--Maruyama (Alg.~\ref{alg:euler_maruyama}) is an algorithm for approximating numerical solutions to SDEs. It is a simple extension of Euler's method, which is the most basic numerical ODE solver, and a technique with which many will be familiar. Like Euler's method, Euler--Maruyama approximates a trajectory by taking small steps tangent to the trajectory. Smaller steps enable approximation with greater precision. The sampling techniques of many common diffusion models~\cite{song2020score, song2020denoising, nichol2021improved} can be viewed as modifications of Euler--Maruyama.

\renewcommand{\algorithmicrequire}{\textbf{Input:}}
\renewcommand{\algorithmicensure}{\textbf{Output:}}

\algdef{SE}[SUBALG]{Indent}{EndIndent}{}{\algorithmicend\ }%
\algtext*{Indent}
\algtext*{EndIndent}

\begin{algorithm}
\caption{Euler--Maruyama Approximation}\label{alg:euler_maruyama}
\begin{algorithmic}[1]
    \Require An SDE of the form $\text{d}\diffusevar = a(\diffusevar, t)\text{d}t + b(\diffusevar, t)\text{d}\qvec{w}$; an initial condition $\diffusevar_0$; a time interval $[0, T]$. A finite number of subintervals $N$.
    \Ensure A simulated trajectory $\{\hat{\diffusevar}_0, \hat{\diffusevar}_1, \hdots,  \hat{\diffusevar}_{N-1}\}$.
    \State \textbf{Initialize:} Partition the time interval $[0, T]$ into $N$ equal sub-intervals $\tau_0 < \tau_1 < ... <\tau_{N}$, where $\tau_{i+1} - \tau_i = \Delta t = T/N$;
    $\hat{\diffusevar}_0 = \diffusevar_0$; n = 0.
    \While{n < N}
        \State $\hat{\diffusevar}_{n+1} = \hat{\diffusevar}_{n} + a(\hat{\diffusevar}_n, \tau_n) \Delta t + b(\hat{\diffusevar}_n, \tau_n) \Delta \qvec{w}_n$,
        \Indent
            \State where $\Delta \qvec{w}_n = \qvec{w}_{\tau_{n+1}} - \qvec{w}_{\tau_n}$
        \EndIndent
        \State $n = n + 1$.
    \EndWhile
\end{algorithmic}
\end{algorithm}

Viewing image synthesis with diffusion models through the lens of numerical SDE solvers can give us intuition about the behavior of different sampling schemes and some insight into how some works have improved the computational efficiency of generating images. Empirically, we observe that generating high-quality images with diffusion models requires many (often hundreds) of iterations; fewer iterations produce poor samples. In the language of numerical differential equation solvers, poor quality results with few iterations is a result of truncation error---the smaller we make our timesteps $\Delta t$, the more accurate our numerical approximation. For the same reason, higher-order differential equation solvers~\cite{karras2022elucidating, dockhorn2022genie} can reduce the error in our numerical approximation, allowing us to sample with greater accuracy or enabling equal quality with fewer network evaluations.

For any diffusion process, there exists a corresponding deterministic process, which can be described by an ODE, that recovers the same marginal probability densities $p(\diffusevar, \sigma)$. Song et al.~\cite{song2020score} define an ODE that describes the deterministic process and name it the probability flow ODE (Eq.~\ref{eq:probability_flow_ode}, Fig. \ref{fig:fundamentals_sdes}b):

\begin{equation}
    \text{d}\diffusevar = \left[\mathbf{f}(\diffusevar, t) - \frac{1}{2}g(t)^2 \nabla_{\diffusevar}\log p(\diffusevar; \sigma(t))\right]\text{d}t.
    \label{eq:probability_flow_ode}
\end{equation}

This probability flow ODE enables deterministic image synthesis---instead of sampling random noise and simulating the reverse-time SDE to synthesize images, we can instead sample random noise and solve the reverse deterministic probability-flow ODE; doing so recovers the same distribution of images. Unlike stochastic sampling, where the final image is determined both by the initial noise image $x_t$ and noise injected at every iteration (corresponding to the $d\qvec{w}$ term in Eq.~\ref{eq:ve_reverse}), an image created deterministically is defined only by the initial noise.

But we can also do the reverse: Instead of sampling noise and producing an image, we can draw an arbitrary image and follow the forward probability flow ODE to encode the image into noise. In fact, the probability flow ODE defines a bijective mapping between images and (noisy) latents. With sufficient accuracy in the ODE solver, one can encode an image into latent space by solving the probability flow ODE forward in time, arrive at a latent $\diffusevar_T$, and solve the probability flow ODE backwards in time to recover the original image. Moreover, one can edit an image by manipulating the corresponding latent. For example, interpolation in latent space may produce a compelling interpolation in image space, and scaling the latent can influence the temperature of the generated image~\cite{song2020score}. Several recent image editing works are built on the premise of a deterministic mapping between images and latents, and rely on the forward probability flow ODE to map real images into the latent space of diffusion models~\cite{hertz2022prompt, mokady2023null, su2022dual}.

Another advantage of deterministic sampling is that the probability flow ODE can also often be solved sufficiently accurately with fewer iterations~\cite{song2020score, song2020denoising} of a numerical solver than the corresponding SDE; as a result, deterministic sampling may produce high-quality images with fewer network evaluations than stochastic sampling, accelerating inference.

Despite these advantages, stochastic sampling is still often preferred over deterministic sampling when many (generally hundreds to thousands) of denoising iterations are available, and image quality is paramount. Intuitively, stochasticity can be seen as a corrective force that repairs errors made earlier in sampling. Thus, stochastic samplers, when combined with many denoising iterations, often produce images that evaluate best according to metrics.

Karras et al.~\cite{karras2022elucidating} illustrate this by decomposing the reverse-time SDE into the sum of a probability flow ODE, which deterministically moves a sample between noise levels, and a Langevin diffusion SDE, which stochastically ``churns'' a sample at a fixed noise level by adding and removing a small amount of noise. Here, the Langevin diffusion SDE, i.e., the stochasticity component, pushes a sample towards the marginal distribution $p(\diffusevar, \sigma)$, correcting errors that may have been carried over in the purely deterministic setting. In practice, deterministic sampling and stochastic sampling have specific strengths and weaknesses, and the level of stochasticity that works best will depend on the task.



\subsection{Latent Diffusion using the Stable Diffusion Model}
Unlike generative models such as Variational Autoencoders (VAEs) or Generative Adversarial Networks (GANs), which generate images through a single forward pass, diffusion models necessitate recurrent forward passes. This characteristic imposes a higher computational burden during training, as the model must learn denoising across multiple noise scales. Additionally, the iterative nature of the multi-stage denoising process elongates inference time, making diffusion models computationally less efficient than their generative counterparts~\cite{karras2022elucidating}.

Generating high-resolution images with diffusion models is often infeasible on consumer-grade GPUs due to the excessive memory requirements. Even on high-end GPUs, the constraints on batch sizes prolong the training process, making it impractical for a large segment of the research community.

\begin{figure}[t]
    \centering
    \includegraphics[width=\columnwidth]{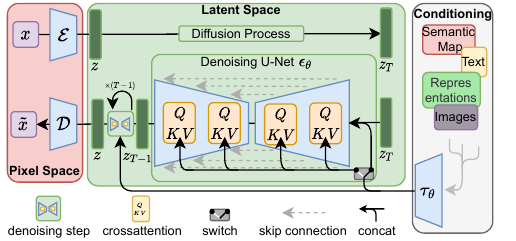}
    \caption{\textbf{Stable Diffusion.} This schematic shows an overview of the latent diffusion approach, including encoder $\mathcal{E}$, decoder $\mathcal{D}$, and conditioning using a cross-attention mechanism. Figure adapted from~\cite{rombach2022high}.}
    \label{fig:latent-diffusion}
\end{figure}

\paragraph*{Perceptual Image Compression.} To address these challenges, Rombach et al.\cite{rombach2022high} introduced latent diffusion models that operate in a compressed latent space, rather than directly on image pixels. This approach retains perceptually relevant details while significantly reducing computational cost. The compressed image space is obtained using an encoder--decoder architecture. Among the various techniques, VQ-GAN\cite{esser2021taming} has emerged as the most common choice due to its impressive compression ability and preservation of perceptual quality. These latent diffusion models (LDMs) consist of a two-stage process: an initial autoencoder for image reconstruction and a subsequent denoising model operating on the latent codes, achieving superior performance with reduced computational demands (see Fig.~\ref{fig:latent-diffusion}).

\paragraph*{Architecture.} The architecture proposed by Rombach et al.\cite{rombach2022high} builds upon the U-Net framework\cite{ronneberger2015u, ho2020denoising, jolicoeurmartineau2020adversarial}. It incorporates attention mechanisms~\cite{vaswani2017attention}, specifically self-attention and cross-attention blocks, at various stages of the U-Net. In the self-attention block, features derived from intermediate U-Net outputs are projected into queries $\mathbf{Q}$, keys $\mathbf{K}$, and values $\mathbf{V}$. The output of the block is given by:
\begin{equation}
    \mathbf{A} \cdot \mathbf{V} \;\; \text{where} \;\; A = \textrm{Attention}(\mathbf{Q}, \mathbf{K}).
\end{equation}
Here, the Attention mechanism captures contextual information between the $d$-dimensional $\mathbf{Q}$ and $\mathbf{V}$ projection matrices via
\begin{equation}
    \textrm{Attention}( \mathbf{Q},\mathbf{K}) = \textrm{Softmax}(\mathbf{Q} \mathbf{K}^T/\sqrt{d}).
    \label{eq:attention}
\end{equation}
Cross-attention blocks operate similarly, enabling controlled generation by injecting conditioning signals such as text prompts (see~Sec.~\ref{sec:fundamental_conditioning_guidance}).

\paragraph*{Retrieval-augmentation Mechanism (RDM).} To further optimize computational efficiency, Blattmann et al.~\cite{blattmann2022retrieval} introduced a Retrieval-augmentation Mechanism (RDM) that fetches relevant image patches from an external database during the generative process. These patches are selected based on latent codes from a pre-trained autoencoder and undergo simple augmentations.
Since details from the retrieved image patches need not be saved in model parameters anymore,
this mechanism leads to a more streamlined denoising model and hence accelerates both training and inference, albeit at the cost of added computational complexity and dependency on a well-trained autoencoder.


\subsection{Conditioning and Guidance}
\label{sec:fundamental_conditioning_guidance}
\paragraph*{Conditioning.}
Perhaps the most important property of a generative model is the ability to control generation through user-defined conditions. Such conditions include text~\cite{reed2016generative}, semantic maps~\cite{park2019semantic}, sketches~\cite{voynov2023sketch}, multi-modal combinations of conditions~\cite{zhan2023mise}, and other image-to-image translation tasks~\cite{isola2018imagetoimage, saharia2022palette}. Formally, instead of sampling data from an unconditional distribution $p(\diffusevar)$, we would like to sample from a conditional distribution $p(\diffusevar|\varcond)$ given some conditioning signal $\varcond$.

To accommodate the variety of conditioning modalities, a flexible set of conditioning mechanisms has been developed for diffusion models. The most simple of these methods is concatenation~\cite{saharia2022palette}, where the condition is directly concatenated with intermediate denoising targets and passed through the score estimator as input. Concatenation can be performed along with the diffusion model input during different stages of the model architecture. It is also applicable to nearly all conditioning modalities. Most notably, Palette~\cite{saharia2022palette} tackles various image-to-image translation tasks such as in-painting, colorization, uncropping and image restoration using conditioning by concatenation.

Another effective method is to inject conditioning signals through cross-attention. Rombach et al.~\cite{rombach2022high} modifies the U-Net architecture~\cite{ronneberger2015u} for conditioning control with cross-attention mechanisms. To control the image synthesis, a conditioning signal $\varcond$, for example a guiding text prompt, is first pre-processed by a domain-specific encoder $\tau$ to an intermediate projection $\tau(\varcond)$. The projected conditioning signal is then injected into the intermediate layers of the denoising U-Net by means of cross attention~\cite{vaswani2017attention}, via Eq.~\ref{eq:attention}, with
\begin{equation}
    \mathbf{Q} = \mathbf{W}_Q \cdot \varphi(\qvec{z}_t), \; \mathbf{K} = \mathbf{W}_K \cdot \tau(\varcond), \; \mathbf{V} = \mathbf{W}_V \cdot \tau(\varcond),
\end{equation}
where $\mathbf{W}_Q, \mathbf{W}_K, \mathbf{W}_V$ are learnable projection matrices, and $\varphi(\qvec{z}_t)$ represents an intermediate result from the denoising U-Net; see Fig.~\ref{fig:latent-diffusion} for a detailed visualization. Intuitively, $\mathbf{Q}$ is the projection of activations of intermediate U-Net layers, while $\mathbf{K}$ and $\mathbf{V}$ are obtained via projection of the given condition.

\begin{figure}[t]
    \centering
    \includegraphics[width=\columnwidth]{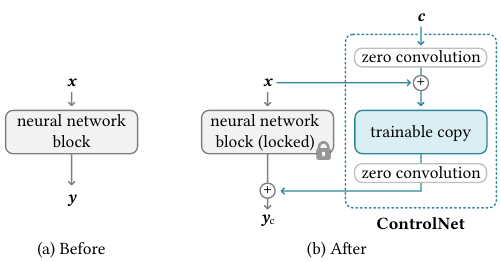}
    \caption{\textbf{Overview of ControlNet.} ControlNet~\cite{zhang2023adding} modifies existing network architectures by duplicating network blocks and connecting them through zero convolutions. The auxiliary module takes some conditioning $\varcond$, allowing the model to learn additional control handles.}
    \label{fig:controlnet}
\end{figure}

During inference, additional techniques can be applied to condition the network on different modalities, including sketches~\cite{voynov2023sketch} and spatial layout~\cite{zhang2023adding,mou2023t2i}.
Among these, adapter methods, exemplified by the popular ControlNet~\cite{zhang2023adding} (see Fig.~\ref{fig:controlnet}), provide an effective and flexible route to add new condition modality without altering the pre-trained diffusion model by embedding new module layers into the existing network architecture.
Specifically, it proposes a backbone for learning diverse control handles for large pre-trained diffusion models through the addition of auxiliary network modules.
New network modules are initialized by duplicating encoding layers from the pre-trained network and connected to the original model through ``zero-convolutions'', a mechanism that initializes layer parameters to zero, ensuring that no harmful noise is learned during the fine-tuning process.
The auxiliary network is fine-tuned on a given set of condition-output pairs, while the original network layers remain unchanged as illustrated in Fig.~\ref{fig:controlnet}.

\paragraph*{Guidance.}

While conditioning affords a level of control over the sampled distribution, it falls short in fine-tuning the strength of the conditioning signal within the model. Guidance emerges as an alternative, generally applied post-training, to more precisely steer the diffusion trajectory.

Dhariwal et al.~\cite{dhariwal2021diffusion} observed that an auxiliary classifier can steer an unconditional generative model. This technique, termed classifier guidance, alters the original diffusion score by incorporating the gradient of the log-likelihood from a pre-trained classifier model \( p_\phi(\varcond|\diffusevar)\) that estimates $\varcond$ from a given image $\diffusevar$. Using Bayes' theorem, the score estimator for $p(\diffusevar|\varcond)$ is given by:
\begin{equation}
\nabla_{\diffusevar_t}\log (p_\theta(\diffusevar_t)p_\phi(\varcond|\diffusevar_t)) = \nabla_{\diffusevar_t}\log p_\theta(\diffusevar_t) + \nabla_{\diffusevar_t}\log p_\phi(\varcond|\diffusevar_t).\label{eq:bayes}
\end{equation}
We can then define a new score estimator $\tilde{\epsilon}(\diffusevar_t)$ corresponding to the joint distribution, giving:
\begin{equation}
\tilde{\epsilon}_\theta(\diffusevar_t, \varcond) = \epsilon_\theta(\diffusevar_t, \varcond) - w\,\sigma_t \nabla_{\diffusevar_t}\log p_\phi(\varcond|\diffusevar_t),\label{eq:classifier_guidance}
\end{equation}
where $w$ is a tunable parameter controlling guidance strength. Despite its versatility, classifier guidance has limitations, such as the need for a noise-robust auxiliary classifier and the risk of poorly defined gradients due to irrelevant information in \(\diffusevar_t\).

To circumvent the limitations of classifier guidance, Ho and Salimans introduced classifier-free guidance~\cite{ho2022classifier}. This method directly alters the training regimen, utilizing a single neural network to represent both an unconditional and a conditional model. These models are jointly trained, with the unconditional model parameterized by a null token \( \varcond = \varnothing \). The model can then be sampled as follows, using the previously introduced
guidance scale \( w \):
\begin{equation}
    \hat{\epsilon}_\theta(\diffusevar_t) = (1+w)\epsilon_\theta(\diffusevar_t, \varcond) - w\epsilon_\theta(\diffusevar_t, \varnothing).
    \label{eq:cfg}
\end{equation}

Controlling the strength of the conditional signal leads to a trade-off between diversity and sample quality~\cite{ho2022classifier}.
As guidance scale $w$ increases, the diversity of the resulting sample decreases in return for higher sample quality.
However, it is often observed that models trained using classifier-free guidance tend to generate low-quality samples for very low or high guidance scales. For example, Stable Diffusion~\cite{rombach2022high} generates empty gray images when sampled using only the unconditional score estimator ($w = -1$), and outputs images with saturation artifacts at higher guidance values ($w > 10$).

\subsection{Editing, Inversion and Customization}
\label{sec:fundamentals:customization}

A pretrained diffusion model essentially provides an expressive generative prior, which can be leveraged to allow average users to perform various image manipulation tasks without any experience on pixel-level crafting skills.
Many recent work have investigated the use of text-to-image diffusion models for \emph{editing}, and generation of personalized images, also known as \emph{customization}.
This section surveys the main works in these two categories, as well as a crucial technical component, \emph{inversion}, which is often used as a building block for editing.

\paragraph*{Editing.}
Owing to its inherently progressive and attention-based architecture, diffusion models offer a unique platform for fine-grained image editing by facilitating adjustments to various network phases and components to manipulate both spatial layout and visual aesthetics. The research trajectory in this domain is geared towards enhancing editing controllability and flexibility, while simultaneously ensuring an intuitive user interface. A prevalent use-case involves altering the visual attributes of an image while retaining its spatial configuration. In this context, SDEdit~\cite{meng2021sdedit} presents a straightforward approach that introduces a calibrated level of noise to an image, resulting in a partially-noised image, followed by a reverse diffusion process with a new conditioning signal as guidance. This foundational approach has been further extended by~\cite{kim2022diffusionclip} to manipulate global characteristics by direct text prompt modification, while localized editing is accomplished through the incorporation of auxiliary masks in the diffusion process~\cite{avrahami2022blended,nichol2021glide}. Similarly, an additional guidance signal ({i.e.}, any gradient function, as a classifier is in classifier guidance) can be used to modify a sampling trajectory to perform manipulative edits like changing object appearances or rearranging content in the scene~\cite{epstein2023diffusion}.
Another editing strategy involves constraining specific feature maps derived from a different generative processes. For example, Prompt-to-Prompt~\cite{hertz2022prompt} employs fixed cross-attention layers to selectively modify image regions corresponding to specific textual cues. Plug-and-Play~\cite{tumanyan2023plug} explores the injection of spatial features and self-attention maps to maintain the overall structural integrity of the image. \cite{cao2023masactrl} advocates for leveraging the self-attention mechanism to enable consistent, non-rigid image editing without the necessity for manual tuning.
Moreover, the text prompt itself serves as a critical determinant of editing quality. Imagic~\cite{kawar2023imagic} refines the text prompts through an optimization of textual embeddings coupled with model fine-tuning, thereby enabling diverse, spatially non-rigid image editing. Delta Denoising Score~\cite{hertz2023delta} ingeniously utilizes the generative prior of Text-to-Image (T2I) diffusion models as a loss term in an optimization framework to guide image transformations based on textual directives. InstructPix2Pix~\cite{brooks2023instructpix2pix} simplifies the user's text input from descriptive target image annotations to more intuitive editing directives by fine-tuning a T2I model on a generated dataset of image pairs aligned with editing instructions, created using a combination of Prompt-to-Prompt and a large language model. \cite{parmar2023zero} obviates the need for text prompts altogether by introducing an automated mechanism for discovering editing directions from exemplar image pairs.

Following similar work in the GAN literature~\cite{pan2023_DragGAN}, a number of diffusion-based methods aim to perform edits driven by sparse user-annotated correspondences, where the appearance and identity of objects are preserved, and only their layout or orientation are changed~\cite{mou2023dragondiffusion,shi2023dragdiffusion}.

\paragraph*{Inversion.}
Many methods for editing an existing ({i.e.}, real) image through generative models often involve an ``inversion'' task, which identifies a specific input latent code or sequence of latents that, when fed into the model, reproduces a given image.
Inversion allows manipulation in the latent space, which enables the use of generative priors already learned by the model.
In the context of diffusion, DDIM inversion \cite{song2020denoising} is a fundamental technique that adds small noise increments to a given image to approximate the corresponding input noise. When running a reverse diffusion using DDIM with this noise, the original image is reproduced. In the case of text-to-image diffusion, when provided with a specific text--image pairing, the DDIM inversion method tends to accumulate small errors, especially with classifier-free guidance \cite{ho2022classifier}. Null-Text Inversion \cite{mokady2023null} compensates for the timestep drift by optimizing the input null-text embedding for each timestep. EDICT \cite{wallace2023edict} achieves precise DDIM inversion using two coupled noise vectors. \cite{wu2022unifying} showcased a DDPM-inversion method, recovering noise vectors for an accurate image reconstruction within the DDPM sampling framework.

Beyond image-to-noise inversion, in the context of text-to-image models, ``textual inversion'' \cite{gal2022image} offers a framework to convert image(s) into token embeddings. The original work proposed converting a concept that appears in a few images into a single token using optimization. Follow-up works also demonstrated the inversion of a single image into a token using an encoder \cite{gal2023encoder}, or converting the concept into a sequence of per-layer tokens to improve the concept's reconstruction~\cite{voynov2023p+}.

\paragraph*{Customization.}
Recent works have extensively explored the customization of T2I diffusion models, i.e., adapt a pretrained diffusion model to generate better outputs for a specific person or object. The pioneering work DreamBooth~\cite{Ruiz2022DreamBoothFT} achieves this by optimizing the network weights to represent a subject shown in a set of images by a unique token. One line of follow-up works has focused on fine-tuning only specific parts of the network. CustomDiffusion~\cite{kumari2023multi} modifies only the cross-attention layers, SVDiff~\cite{han2023svdiff} refines the singular values of weights, LoRA~\cite{hu2021lora} targets optimizing low-rank approximations of weight residuals~\cite{hu2021lora} and StyleDrop~\cite{sohn2023styledrop} adopts adapter tuning \cite{houlsby2019parameterefficient} to fine-tune a selected set of adapter weights for style customization. Similar techniques have been applied to other problem statements besides text-to-image generation, such as image inpainting or outpainting~\cite{tang2023realfill}.

Another line of research is dedicated to accelerating the customization process. \cite{gal2023designing} and~\cite{wei2023elite} employ encoders to determine initial text embeddings and subsequently fine-tune them to enhance subject fidelity. \cite{ruiz2023hyperdreambooth} predicts low-rank network residuals tailored to specific subjects directly. SuTI~\cite{chen2023subject} initially constructs a comprehensive paired dataset using images and their recontextualized counterparts produced by DreamBooth, and uses it to train a network that can execute personalized image generation in a feed-forward manner. InstantBooth~\cite{shi2023instantbooth} and Taming Encoder~\cite{jia2023taming} introduce a conditional branch to the diffusion model which allows conditioning using a minimal set of images or even just one, facilitating the generation of personalized outputs in various styles. Break-A-Scene~\cite{avrahami2023break} customizes a model to support a few subjects depicted in a single image, while FastComposer~\cite{xiao2023fastcomposer} leverages an image encoder to project subject-specific embeddings for multi-subject generation.

\section{The Challenge of Moving Beyond Images}\label{sec:challenges}

Diffusion models have garnered significant attention and success in the realm of image processing, owing to a confluence of factors that have made them particularly well-suited for this domain. One of the primary reasons for their efficacy lies in the maturity of network architectures tailored for image processing, particularly in the area of denoising.
Diffusion models in the 2D image domain have capitalized on these advancements, incorporating well-defined building blocks such as convolutional layers, attention mechanism and U-Net structures as their backbone. Advances in Transformers~\cite{vaswani2017attention} and large language models~\cite{devlin2018bert} have further enhanced these models by enabling approximate controllability through natural language prompts, facilitated by the pairing of images with text descriptions~\cite{radford2021learning}.

Moreover, the ubiquity of mobile phones and social norms have democratized the capture, storage, and sharing of 2D images, resulting in a near-infinite supply of freely available images. 
In summary, by the end of the 2010s, all the essential ingredients for the success of 2D diffusion models were in place: a well-defined mathematical framework, flexible function approximators capable of learning expressive image transformations, and an abundant supply of training data.

Though 2D image synthesis has been blessed by a happy coincidence of technological progress and available data, this has not been the case for higher-dimensional signals.
The task of synthesizing higher-dimensional content, such as video and 3D content, is substantially more difficult than 2D image synthesis, limited by a number of additional issues described further in this section.

\paragraph*{Models.}
The network architecture for processing high-dimensional data is still an open question. Unlike images, which can be efficiently represented and processed using discrete pixel values, higher-dimensional data often require more complex representations. This complexity is exacerbated by the need to handle long-range information flow, which is critical for understanding the temporal dynamics in videos or the spatial relationships in 3D structures. As of now, there is no consensus on a network architecture that can serve as a reliable and scalable backbone for diffusion models in these domains. The computational and memory costs associated with processing high-dimensional data further complicate the issue, making it challenging to find an efficient yet expressive solution.

\paragraph*{Data.}
The availability and quality of data pose another significant challenge. For 3D structures, the process of creating a single 3D model involves multiple steps, including scanning, processing, and reconstruction, each of which requires specialized expertise and resources. This makes the data acquisition process both time consuming and expensive.
In the case of videos---although raw data may be abundant---annotating these data, especially for capturing motion and temporal dependencies, is far from trivial. This scarcity of high-quality annotated data hampers the training of robust and generalizable diffusion models for higher-dimensional domains.
\section{Video Generation and Editing}\label{sec:video}

Despite the tremendous progress in \emph{image} diffusion models, and the remarkable breakthroughs in T2I generation, expanding this progress to the domain of  \emph{video} is still in nascent stages, due to two main challenges.

First, learning from videos requires orders of magnitude more data than images, due to the complexity and diversity of our dynamic world. While online videos are abundant, curating high-quality video datasets remains a difficult task that typically involves significant engineering efforts and requires dedicated automatic curation tools. 

Another substantial challenge arises from the high dimensionality of raw video data (e.g., a two-minute video with 30~fps consists of 3600$\times$ more pixels than a single frame). This makes the extension of 2D architectures to the temporal domain very challenging and computationally expensive. We next discuss how these challenges have been tackled in the context of diffusion models for video generation. 


\begin{figure}[t!]
    \centering
    \includegraphics[width=\columnwidth]{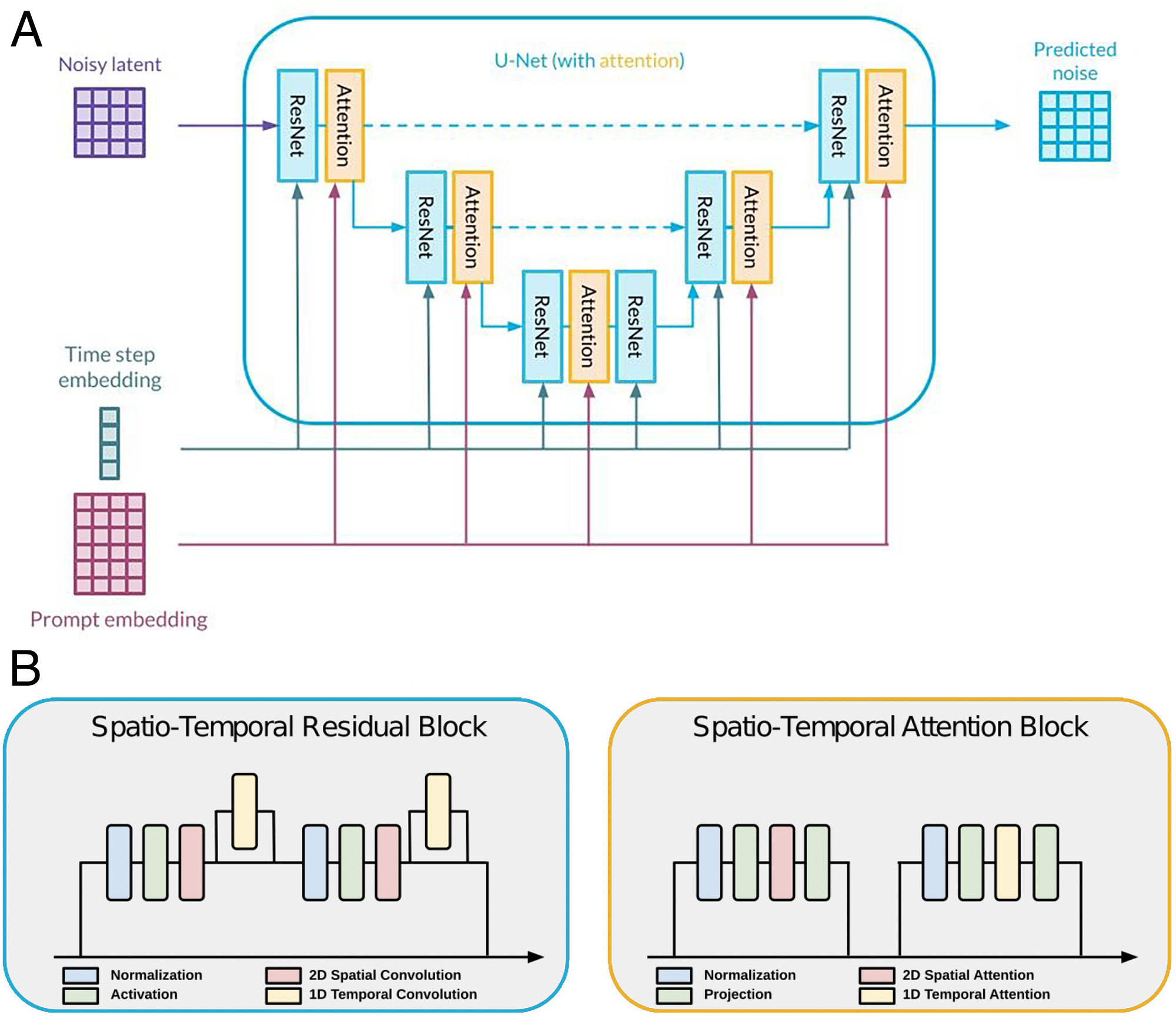}
    \caption{\textbf{Denoising U-Net with Spatio-temporal Attention.} (A) Common attention U-Net used for the denoising step in image and video diffusion models comprising residual convolution blocks and attention blocks as well as concatenated time step and text prompt--embeddings. Figure adapted from \cite{Kochanowicz}. (B) Temporal structure in video diffusion and editing is commonly modeled by adding 1D temporal convolutions in the residual blocks (left) as well as 1D temporal attention blocks after each of the 2D spatial attention blocks (right). Figure adapted from~\cite{esser2023structure}.}
    \label{fig:video:spatiotemporalunet}
\end{figure}

\subsection{Unconditional and Text-Conditioned Video Diffusion}
There have been significant research efforts in extending diffusion models to the temporal domain, aiming to capture the vast distribution of natural motion  from a large-scale video dataset.
Ho et al.~\cite{ho2022video} introduced the first Video Diffusion Model (VDM), extending the 2D U-Net backbone to the temporal domain. This is achieved through  factorized space and time modules, enabling more efficient computation as well as joint training on both individual images, videos, and text. This approach has been scaled up by Imagen Video~\cite{ho2022imagen}, a cascaded text-to-video (T2V) model with 11 billion parameters, which comprises a low spatio-temporal resolution base model, followed by multiple cascaded super-resolution models that increase both the spatial resolution and the effective framerate.  Imagen Video is \emph{trained from scratch}, using a large corpus of high-quality video and corresponding captions as well as a number of text-image datasets.

Aiming to re-use learned image priors for video generation, Make-A-Video~\cite{singer2022make} builds their framework on a pre-trained T2I model, extending it to videos by adding spatio-temporal convolution and attention layers to the existing T2I 
model, followed by spatial and temporal super-resolution models.  A key property of this approach is that each component can be trained separately:  The T2I model is pre-trained over image-text pairs, while the entire T2V is fine-tuned on a large-scale corpus of \emph{unlabeled} videos, thus bypassing the need to have video--caption-paired training data. This effectiveness of inflating and finetuning a T2I model for video generation has been also demonstrated  in autoregressive Transformer-based models (e.g.,~\cite{hong2022cogvideo}).

These two components---using factorized/separable spatio-temporal modules (Fig.~\ref{fig:video:spatiotemporalunet}) and building upon a pre-trained text-to-image model---have been also used to expand an image Latent Diffusion Model (LDM), e.g., Stable Diffusion~\cite{rombach2022high}, to videos, i.e., learning the video distribution in a low-dimensional space (e.g., \cite{blattmann2023align,li2023videogen,zhou2022magicvideo,wang2023modelscope}). Here too, compared to vanilla Video Latent Models (e.g., \cite{he2022latent,yu2023video}),  leveraging a pre-trained image model allows efficient training (with less data), while harnessing the rich 2D priors learned by T2I LDMs. Another advantage of this approach is the ability to transfer the learned motion modules to other image models derived from the
same base T2I. For example,  plugging in a personalized tuned version of the T2I model thus synthesizing videos in a specific style or depicting specific objects \cite{blattmann2023align,guo2023animatediff}. 

As illustrated in Fig.~\ref{fig:video:spatiotemporalunet}, the factorized spatio-temporal modules expand both the convolutional block by a 1D convolution in time and also the self-attention block to model dynamics. A common approach to ``inflating'' self-attention (see Sec.~\ref{sec:fundamentals}) to multiple frames, aka extended or cross-frame attention, expands the self-attention across all or a subset of the frames of a video as 
\begin{align}
 \textrm{Softmax} \left( \frac{\mathbf{Q}^i \left[ \mathbf{K}^1, \ldots, \mathbf{K}^I \right]^T}{\sqrt{d}} \right) \cdot \left[ \mathbf{V}^1, \ldots, \mathbf{V}^I \right],
\end{align}
where $\mathbf{Q}^i, \mathbf{K}^i, \mathbf{V}^i$ are the queries, keys, and values of frame $i=\{1 \ldots I\}$. This inflation mechanism is prevalent among many methods discussed in this section. While it promotes temporal consistency, it does not guarantee it.

Other approaches for video generation include MCVD~\cite{voleti2022MCVD}, which autoregressively generates videos in a 3D latent space  by conditioning new frames on previously generated ones; VideoFusion~\cite{luo2023videofusion}, which decomposes the noisy video latents into the sum of a (static) base noise shared across all frames, and a (dynamic) residual per-frame noise; 
and Generative Image Dynamics~\cite{li2023generativeimagedynamics}, which instead of directly predicting video content, learns to generate motion trajectories for pixels in an image, such that images can be animated to arbitrary length videos with oscillating motion.

\begin{figure}
    \centering
    \includegraphics[width=\columnwidth]{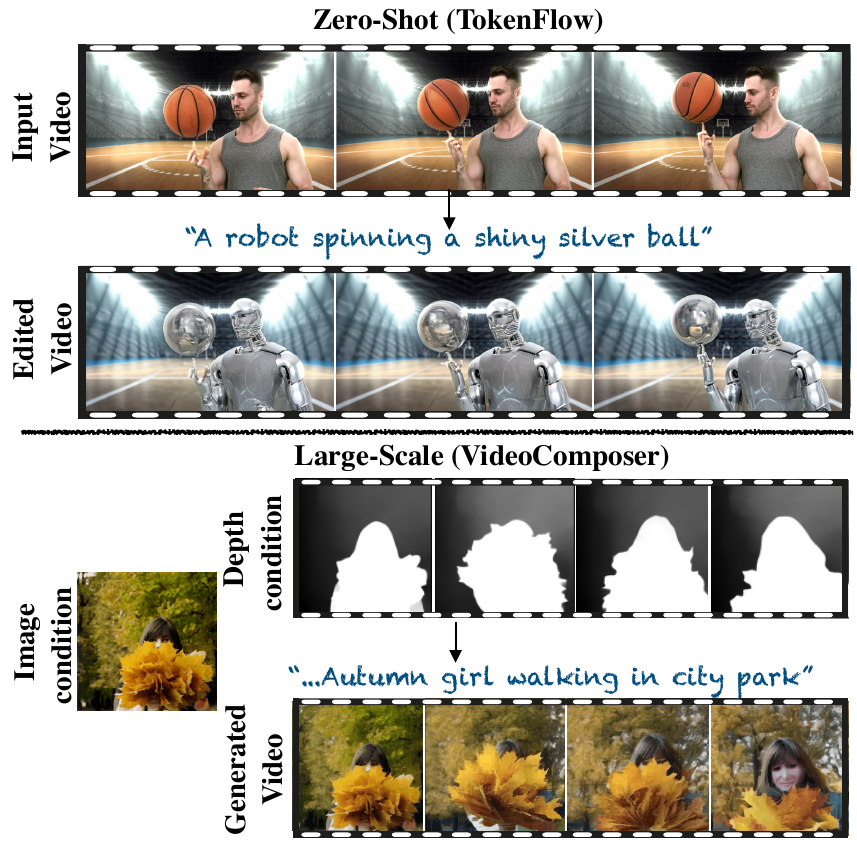}
    \caption{\textbf{Text-driven video editing.} Top: Given an input video, TokenFlow~\cite{tokenflow2023}---a zero-shot method that leverages a pretrained T2I model---enables consistent editing according to a given text prompt. Bottom: VideoComposer~\cite{wang2023videocomposer}---a conditional video diffusion model trained on a large-scale video dataset---enables various controls, including conditioning video generation on a given image and per-frame depth maps.}
    \label{fig:vid-editing}
\end{figure}
\subsection{ Controlled Video Generation and Editing}\label{ssec:video_generation}
Similar to images, an important aspect in harnessing diffusion models for real-world content creation tasks, is the ability to provide users with controls over various attributes of the generated content, ranging from texture/appearance to editing motion and actions in video.  While powerful methods have been developed to control various  image attributes in T2I models,  video editing poses additional challenges. First, any edit has to be applied in a consistent manner to all video frames. Second, while the community has developed rich and powerful representations for ``static image attributes'',  the question of how to represent motion and time-varying signals in videos still remains open.

\paragraph*{Conditional Video Models.}
One approach for controlled video generation is to design and train a conditional video model that directly takes  control signals as input.
Most existing works have been focused on video-to-video translation, where  the overall motion and layout are extracted from a driving video. For instance, Runway's Gen-1 model~\cite{esser2023structure} takes as input per-frame depth maps, which control the spatial layout of each frame, and a CLIP embedding, which controls the global appearance and semantic content of the video. Control-A-Video~\cite{chen2023controlavideo} and VideoComposer~\cite{wang2023videocomposer} enable video generation conditioned on various other image-space control signals, such as edge maps, sketches or pose, by extending ideas from controlled image generation to the realm of videos \cite{zhang2023adding,huang2023composer}.
All the above models are based on inflated T2I models, which typically incorporate additional temporal layers, and are trained on large-scale video datasets. Similar concepts have been explored for more limited domain videos, for example models aimed at animating videos of humans, conditioned on an input image and a sequence of target poses~\cite{karras2023dreampose}.
MagicEdit~\cite{liew2023magicedit} proposes a modular framework, which only trains the added temporal layers and not the pre-trained T2I model. Similar to AnimateDiff~\cite{guo2023animatediff}, this enables plugging the trained temporal layers into a given conditional T2I model (e.g., ControlNet~\cite{zhang2023adding}) for controllable video generation.



\paragraph*{Few-shot Methods for Video Editing.}
On the other side of the spectrum, a surge of methods suggest to leverage a pre-trained T2I model in a  one-shot setting, i.e., by fine-tuning on a single test video (e.g, ~\cite{wu2022tune,liu2023videop2p}), or in a zero-shot setting, i.e., given no additional training data (e.g., \cite{qi2023fatezero,ceylan2023pix2video,khachatryan2023text2video,tokenflow2023,Fridman2023SceneScapeTC}).

For instance, Tune-A-Video \cite{wu2022tune} observes that simply extending the spatial
self-attention in the T2I model from one image to multiple
images produces consistent content and appearance across the generated images. Based on this finding, they design an inflated version of the T2I model, and finetune it on the test video. At inference, given a text prompt, the model can be used to change the object category or stylize the video.

While tuning a T2I model on a single video yields surprisingly promising results, it is prone to overfitting (i.e., forgetting the T2I prior) and demands costly computation, limiting its use to generating text-driven variations of short, sub-sampled clips. Alternatively, a surge of video editing methods leverage a T2I model in a zero-shot fashion, by \emph{directly manipulating its internal features}.  Following feature-based image editing methods~\cite{tumanyan2023plug,hertz2022prompt}, Fate-Zero \cite{qi2023fatezero} extracts the attention maps from an input video, and injects them into the T2I model during the generation of the edited video; this operation allows them to preserve the spatial layout of each frame, as discussed in Sec.~\ref{sec:fundamentals}; appearance consistency  is achieved by extending the self-attention to operate on multiple frames, as in \cite{wu2022tune}. Concurrent works \cite{ceylan2023pix2video,khachatryan2023text2video} combine the self-attention extension with  a conditional T2I (e.g, ControlNet \cite{zhang2023adding}), thus allowing to condition the generation on various spatial controls.   However, achieving highly consistent edits remains challenging, as it is only implicitly encouraged via the inter-frame attention module.

Re-Render A Video \cite{yang2023rerender} aims to improve consistency via a two-step approach by editing keyframes, and applying off-the-shelf video propagation methods to apply the edit on the rest of the frames. Their method heavily relies on accurate optical flow in both keyframe editing and propagation, which limits their use to rather simple motion and short clips. Concurrently, TokenFlow \cite{tokenflow2023} observes that consistency in RGB space is directly affected by the consistency of the  diffusion features across frames. Thus, their  method  ensures temporal consistency by explicitly preserving the inter-frame feature correspondences of the original video. Notably, TokenFlow operates entirely in the diffusion feature space, thus does not exhibit optical-flow restrictions.

\paragraph*{Neural Video Representations for Consistent Synthesis.}

Videos inherently contain highly redundant information across time, i.e., often share the same objects, scene background and overall appearance across frames.  Thus, synthesizing or editing video content in a frame-by-frame manner is a tedious process that is susceptible to introducing temporal inconsistencies.  This has motivated the development of \emph{neural video representations} that facilitate effective and consistent synthesis and editing. Such representations range from video-to-layered-image decomposition to full 4D dynamic scene representations. 

For example, representations like  Layered Neural Atlases (LNA)\cite{kasten2021layered} and Deformable Sprits~\cite{ye2022sprites} decompose a video into a set of  layered canonical images (atlases),  with corresponding per-frame deformation fields. This approach allows to significantly simplify video editing: 2D edits applied to a single canonical image are automatically propagated to all video frames. Text2LIVE~\cite{bar2022text2live} leverages LNA representation in conjunction with CLIP to perform text-driven video editing.  This approach has been extended in \cite{lee2023shape} to leverage a pre-trained T2I model and to enable shape deformations.    While temporal consistency is guaranteed by design, these methods typically require heavy optimization (hours of training per video).  To tackle this issue, CoDeF~\cite{ouyang2023codef} proposes an efficient hash-based representation to both canonical images and deformation fields. 

To ensure both temporal and geometric consistency, 
Make-a-Video3D \cite{Singer2023TextTo4DDS} synthesize a video by leveraging a full 4D video representation, i.e., a 4D dynamic NeRF \cite{cao2023hexplane}, representing the 3D position of each scene point, and its 3D motion throughout the video. Given a target text prompt, this representation is optimized using  a score-distillation-based objective that combines 2D image priors learned by a pre-trained T2I, and motion priors learned by a pre-trained T2V  (see more details in Sec.~\ref{sec:4D_generation}).

\section{3D Diffusion} \label{sec:spatial}
Beyond videos, the advent of diffusion models has also ushered in a transformative era for visual computing in the domain of 3D content generation.
Although videos can be seen as 3D data (2D frames stacked sequentially in time), here we use the term ``3D'' to refer to spatial structure, i.e., 3D geometry.
While 3D scenes can be visually presented as 2D video content by rendering a scene from multiple viewpoints, the rendered videos differ from general video content in that they only contain camera motion (as long as there exists a single 3D-consistent scene geometry).
Generating this scene geometry (and the accompanying appearance or texture) is the primary focus of 3D generative tasks.

As elaborated in Sec.~\ref{sec:challenges}, the application of diffusion models to higher-dimensional data faces inherent difficulties, with data scarcity being particularly acute in the 3D domain.
Existing 3D datasets are not only orders of magnitude smaller than their 2D counterparts, but they also exhibit wide variations in quality and complexity (see Sec.~\ref{sec:data}). Many of the available datasets focus on individual objects with limited detail and texture, limiting their generalization and usefulness for practical applications. Fundamentally, this issue stems from the fact that 3D content cannot be captured (or created) with the same ease as an image or video, resulting in orders of magnitude less total data. Furthermore, the culture for sharing 3D content is nowhere near that for 2D, exacerbating the scarcity of annotated data.

A unique challenge specific to 3D diffusion is the lack of a standardized 3D data representation.
Unlike in image and video processing, where consensus has emerged around certain data formats, the 3D field is still wrestling with multiple competing representations, i.e.,~meshes, points, voxels, and implicit functions, each with its own merits and limitations, and each necessitating different network designs. Yet, none has emerged as a definitive solution that balances scalability with granularity.

To navigate these complexities, this section considers two distinct approaches: 1) \emph{Direct 3D Generation via Diffusion Models}, which aims to model the distribution of 3D shapes, such that 3D content can be generated directly, putting geometry at the forefront and potentially serving as a foundation for tasks like 3D reconstruction and shape retrieval; and 2) \emph{Multiview 2D-to-3D Generation via Diffusion Models}, which offers a more practical route by leveraging high-quality 2D generative models to create textured, consumer-ready 3D content.
Both approaches offer valuable insights and capabilities, each addressing different facets of the challenges and opportunities in 3D content generation via diffusion models.

\begin{table*}[tbhp]
\centering
\setlength{\tabcolsep}{2pt}
\adjustbox{max width=\textwidth}{
\begin{tabular}{cccccHcccHcc}
\toprule
Output & Method & 3D Repr. & \makecell{Diffusion\\Repr.} & \makecell{Latent\\Struct.} & Latent Arch. & \makecell{Diffusion\\Arch.} & \makecell{Super-\\vision} & \makecell{Hierar-\\chical} & Locality & \makecell{Optional\\Conditioning} & Data \\\midrule
\multirow{10}{*}{\rotatebox{90}{object geom.}}& DPM~\cite{luo_diffusion_2021} & points & points & - & NA & PointNet & 3D & \xmark & \xmark & NA & ShapeNet\\
& PVD~\cite{zhou_3d_2021} & points & points       & - & NA & PVCNN  & 3D & \xmark & \xmark & d & ShapeNet\\
& NeuralWavelet~\cite{hui_neural_2022} & TSDF grid & wavelet coefficients & - & NA & 3D U-Net  & 3D & \cmark & \cmark & NA & ShapeNet\\
& LAS-Diffusion~\cite{zheng_locally_2023} & \makecell{Occ. \& SDF grid} & \makecell{Occ. grid \& SDF octree} & - & NA & 3D U-Net  & 3D & \cmark & \cmark & k, c & ShapeNet \\
& LION~\cite{zeng_lion_2022} & points & latents & points & PVCNN & PVCNN  & 3D & \cmark & \cmark & s & ShapeNet \\
& SDFusion~\cite{cheng_sdfusion_2023} & TSDF grid & latents & voxel & 3D U-Net & 3D U-Net  & 3D & \xmark & \cmark & \makecell{s, i, t} & \makecell{ShapeNet, BuildingNet,\\ Pix3D, Text2shape}\\
& HyperDiffusion~\cite{erkocc2023hyperdiffusion} & SDF & net weights & 1D &  NA & Transformer  & 3D & \xmark & \xmark & NA & ShapeNet \\
& Diffusion-SDF~\cite{li_diffusion-sdf_2023} & TSDF grid & latents & voxel & NA & 3D U-Net  & 3D & \xmark & \cmark & \makecell{s, i, t} & \makecell{ShapeNet, Text2shape}\\
& NFD~\cite{shue_3d_2023} & Occ. & latents & triplane             & NA & 2D U-Net  & 3D & \xmark & \xmark & NA & ShapeNet \\
& 3DShape2VecSet~\cite{zhang_3dshape2vecset_2023} & SDF & latents & set  & Perceiver & 1D U-Net  & 3D & \xmark & \cmark & \makecell{s, c, i, t} & ShapeNet, ShapeGlot\\
& Michelangelo~\cite{zhao2023michelangelo} & occupancy & latents & set & Perceiver & 1D U-Net  & 3D & \xmark & \xmark & i, t & \makecell{ShapeNet\\3D Cartoon Monster (not public)} \\\midrule
\multirow{8}{*}{\rotatebox{90}{\makecell{object geom.+ appear.}}}& Point-E\cite{nichol_point-e_2022} & colored points & latents & points  & Transformer & Transformer  & 3D & \cmark & \xmark & t & \textcolor{red}{proprietary} \\
& Shap-E\cite{jun2023shap} & radiance field & net weights   &  1D     & Transformer & Transformer  & 3D & \xmark & \xmark & t & \textcolor{red}{proprietary} \\
& 3DGen~\cite{gupta_3dgen_2023} & textured mesh & latents      & triplane & U-Net & U-Net  & 3D & \xmark & \cmark & i, t & ShapeNet, Objaverse\\
& DiffRF~\cite{muller2023diffrf} & radiance field & latents & voxel &  & U-Net & 3D U-Net & \xmark & \xmark & i & \makecell{Photoshape Chairs, ABO} \\
& RenderDiffusion~\cite{anciukevivcius2023renderdiffusion} & radiance field & latents & triplane & ResNet & U-Net & 2D & \xmark & \xmark & i & \makecell{FFHQ, AFHQv2, \\CLEVR, ShapeNet} \\
& HoloDiffusion~\cite{karnewar2023holofusion} & radiance field & latents & voxel & ResNet & 3D U-Net & 2D & \xmark & \xmark & i & CO3Dv2 \\
& HoloFusion~\cite{karnewar2023holodiffusion} & radiance field & latents & voxel & ResNet & 3D U-Net & 2D & \cmark & \xmark & i & CO3Dv2 \\
& SSDNeRF~\cite{chen2023single} & radiance field & latents & triplane & auto-decoder & U-Net & 2D & \xmark & \xmark & i & ShapeNet, ABO \\\midrule
\multirow{4}{*}{\rotatebox{90}{\makecell{scene \\ geom. + appear.}}}\\
& GAUDI~\cite{bautista2022gaudi} & radiance field & latents  & 1D & MLP, CNN & 1D U-Net & 2D & \xmark & \xmark & i, t, c & \makecell{VizDoom, Replica,\\ VLN-CE, ARKitScenes} \\
& NF-LDM~\cite{kim2023neuralfield} & radiance field & latents & hybrid & 2D U-Net & 1-2D U-Net & 2+3D & \cmark & \cmark & m & \makecell{VizDoom, Replica,\\ Carla, AVD (not public)} \\\\
\bottomrule
\end{tabular}
}
\caption{\textbf{Geometry Generation with Diffusion Models.} We divide the table into three sections corresponding to the generation of object-level geometry, object-level geometry and appearance, and scene-level geometry and appearance.
The conditioning column uses t (text), i (image), d (depth map), k (sketch), m (segmentation map), c (category), s (partial or coarse shape) and NA (not applicable).}
\label{tab:geometry_gen_summary}
\end{table*}

\begin{figure*}[tbhp]
    \centering
    \begin{subfigure}{.25\textwidth}
		\includegraphics[height=4.5cm]{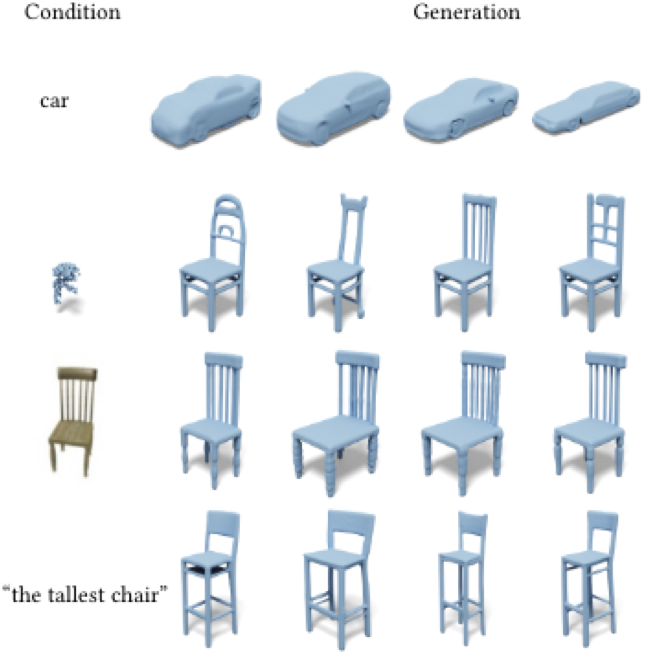}
		\caption{Object-level geometry}\label{fig:object-geometry}
	\end{subfigure}%
	\begin{subfigure}{.28\textwidth}
		\includegraphics[height=4.2cm]{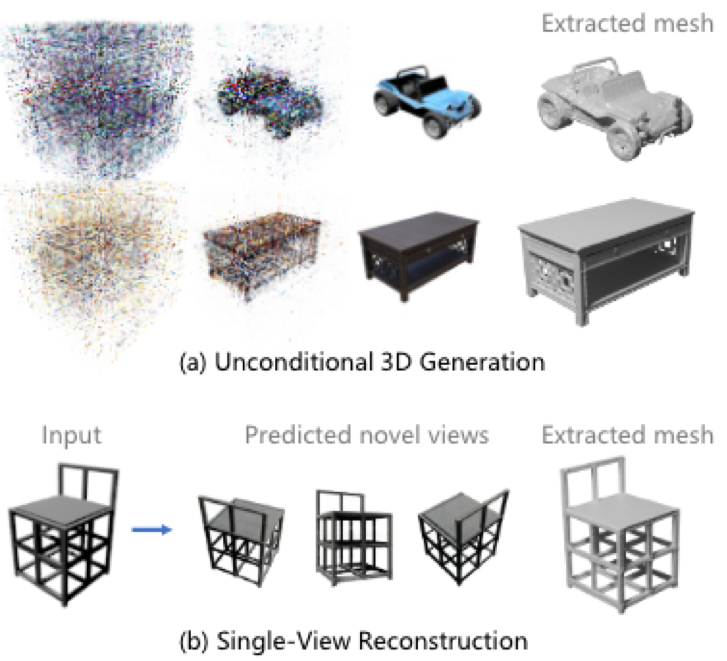}
		\caption{Object-level geometry and appearance}\label{fig:object-geometry-appearance}
	\end{subfigure}%
	\begin{subfigure}{.45\textwidth}
		\includegraphics[width=\linewidth]{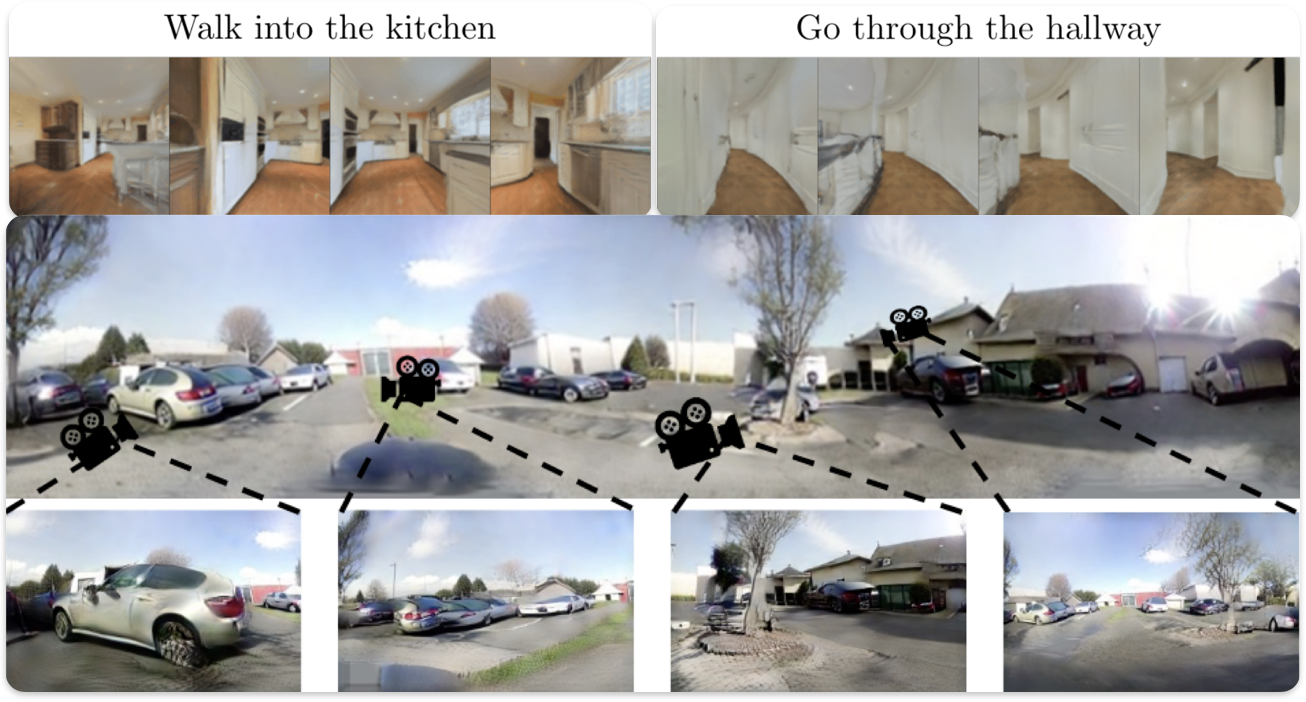}
		\caption{Scene-level geometry and appearance}\label{fig:scene-geometry-appearance}
	\end{subfigure}
    \caption{\textbf{Direct 3D Generation.} Representative examples of direct 3D generation via diffusion models. The examples from left to right depict the state-of-the-art results of the generation of object-level geometry~\cite{zhang_3dshape2vecset_2023}, object-level geometry and appearance~\cite{chen2023single}, and scene-level geometry and appearance generation~\cite{bautista2022gaudi,kim2023neuralfield}.}
    \label{fig:example-direct-3D}
\end{figure*}

\subsection{Direct 3D Generation via Diffusion Models}\label{sec:direct-3d-generation}

Due to the aforementioned challenges inherent to 3D data representation and spatial reasoning, in the realm of ``Direct 3D Generation via Diffusion Models'', the design space is notably intricate, necessitating a nuanced exploration of various design factors that significantly distinguish existing methodologies.
This has led to a diverse array of design choices, each with its own merits and limitations. The ensuing discussion will systematically explore these pivotal design vectors, elucidating their impact on the quality and applicability of generated 3D content.
\Cref{tab:geometry_gen_summary} provides a comprehensive summary of the methods reviewed herein.

\paragraph*{Type of Output.}
The first way to distinguish different methods is to look at the type of output they generate.
Current methods either generate object-level geometry, object-level geometry and appearance, or scene-level geometry and appearance (see Fig.~\ref{fig:example-direct-3D} for an example of the output).
The choice of output type is often dictated by the type of data that is available for training.
Many works~\cite{luo_diffusion_2021,zhou_3d_2021,hui_neural_2022,zheng_locally_2023,zeng_lion_2022,cheng_sdfusion_2023,erkocc2023hyperdiffusion,li_diffusion-sdf_2023,shue_3d_2023,zhang_3dshape2vecset_2023} make use of the few existing large-scale 3D datasets, such as ShapeNet~\cite{chang2015shapenet}, which include object-scale synthetic 3D models with geometry and material provided.
These works use these datasets to investigate various design choices to adapt 2D diffusion models to 3D content; they have demonstrated the potential of diffusion models for 3D generation
with satisfying geometric details---as the training data permits.
However, due to the limited size, diversity, and complexity of the data, the applicability of these methods in practice is relatively low.
Especially the lack of diversity and appearance information is a major limitation.
To address this, several works, including \cite{nichol_point-e_2022,jun2023shap,gupta_3dgen_2023,zhao2023michelangelo}, have started to use more complex datasets, most notably~\cite{reizenstein2021common,objaverse,ObjaverseXL} with in-the-wild objects and their appearance.

More recently, several authors have started to explore the generation of large indoor and outdoor 3D scenes.
The scale and complexity of these scenes are much higher than the object-level datasets, though, the diversity is still limited.
Interesting, likely due to the lack of accurate 3D ground truth and/or suitable architectures to process large-scale 3D structures, existing scene-level generation methods~\cite{bautista2022gaudi,kim2023neuralfield} opt to generate video renderings of the scene, which includes both the static 3D scene and the camera trajectory.
While this is seemingly harder than generating the 3D scene only, due to the additional output mode, these approaches can conveniently hide low-quality geometry through visually plausible textures and appearance.
Normally, the generated 3D scenes are coarse and incomplete, which leads to uncanny warping effects in the generated video, creating an illusion that the shape is deforming over time.
In addition, the generated scenes are still relatively simple and lack the diversity and complexity of real-world scenes.

\paragraph*{3D Shape Representation.}
Given the type of output, there can be multiple suitable representations, for example, point clouds, voxels, meshes, continuous or discrete samples of signed distance functions (SDFs), occupancy, and  radiance fields.
For object-level geometry, points, meshes, occupancy and SDFs---both as continuous functions or discrete samples---are common choices.
When appearance is also generated, points and meshes can be augmented with per-point colors~\cite{nichol_point-e_2022} and textures~\cite{gupta_3dgen_2023}, and the occupancy and signed distance function can also be augmented with texture field to form some variations of radiance field~\cite{jun2023shap,zhao2023michelangelo}.
For scene-scale outputs, due to the lack of high-quality data in points and meshes representations and relatively easier acquisition in RGBD format, radiance fields become a more viable choice.

\paragraph*{Diffusion Space.}
One very important way to differentiate the existing frameworks is based on the representation they employ inside the diffusion process.
A natural choice is to apply the diffusion process directly to the 3D shape representation.
The most popular representation in this category is points~\cite{luo_diffusion_2021,zhou_3d_2021} and voxelized occupancy and SDF~\cite{zheng_locally_2023}, as the former can be conveniently linked to the physics interpretation of diffusion models with Langevin dynamics and the latter, deploying voxel structures, can take advantage of tested 2D diffusion architectures by simply adding a spatial dimension to all network operations.
However, this representation simply captures raw and uncompressed information, it is very inefficient in terms of memory and computation and thus fails to capture high-frequency details with a fixed spatial resolution.
A popular alternative which addresses this issue is applying the diffusion process on a more expressive and compact (latent) representation, extracted from the 3D data through some transformation.
The transformation can be deterministic such as a wavelet transform used in Neural Wavelet~\cite{hui_neural_2022}, or learned as a high-dimensional latent code~\cite{zeng_lion_2022,zhou_3d_2021,li_diffusion-sdf_2023,shue_3d_2023,zhang_3dshape2vecset_2023,nichol_point-e_2022,jun2023shap,gupta_3dgen_2023,zhao2023michelangelo, bautista2022gaudi, kim2023neuralfield} or the weights of the trainable neural implicit representation themselves, for example neural occupancy fields or SDFs, such as in~\cite{erkocc2023hyperdiffusion,jun2023shap}.
These methods typically consist of two steps: First, a piecewise constant or smooth latent code is learned from the 3D data;
Subsequently, a diffusion model is trained to denoise these latent codes, while the autoencoder model remains fixed.
The smoothness of the latent space is crucial for the success of this approach.
It is often achieved by imposing a regularization term on the latent codes, such as the total variation loss~\cite{yu2022plenoxels,shue_3d_2023}, or deploying a variational framework such as (vector quantized) variational autoencoder~\cite{autosdf2022}.
While the two-step approach has been predominant, a series of concurrently developed works~\cite{anciukevivcius2023renderdiffusion,karnewar2023holodiffusion,karnewar2023holofusion,chen2023single} demonstrate how these two steps can be unified into end-to-end training, which is not only more efficient in a practical sense but also mutually beneficial for the two steps.
We will discuss this in greater detail in the \textbf{Types of Supervision} paragraph.

\paragraph*{Structure of the Latent Codes.}
While the simplest form of latent code takes the form of a 1D vector~\cite{erkocc2023hyperdiffusion,jun2023shap,bautista2022gaudi}, most approaches opt for a more spatially aligned latent code to improve the locality and capability of the latent representation.
This structure governs computational efficiency and memory requirements and, more importantly, it is coupled with the choice of diffusion architecture.
In fact, the popularity of well-established convolution-based diffusion architectures has led to the widespread use of voxel grids~\cite{cheng_sdfusion_2023,li_diffusion-sdf_2023,karnewar2023holodiffusion,karnewar2023holofusion}.
Still using convolution, the triplane structure~\cite{chan2022efficient} is also a popular choice\cite{shue_3d_2023,gupta_3dgen_2023,anciukevivcius2023renderdiffusion,wang_rodin_2023}, which factorizes a 3D volume to three axis-aligned orthogonal planes~\cite{peng2020convolutional, chan2022efficient} and thereby significantly reduces the memory and computation cost.

In an alternative approach, some methods have adopted point clouds as the data structure for latent codes~\cite{zeng_lion_2022, nichol_point-e_2022}.
In this paradigm, each latent code is associated with a specific point in the 3D space, thereby enabling localized information storage.
The spatial locations of these points are either learned or inferred from the input shape during the initial latent learning phase.
In a more abstract vein, 3DShape2VecSet~\cite{zhang_3dshape2vecset_2023} eschews spatial information altogether by eliminating the coordinates from the latent codes.
Consequently, the latent codes become a set of unbounded codes, offering greater flexibility to model long-range dependencies and self-similarities at the expense of spatial locality.
These models often employ architectures like common point-processing networks such as transformers and point-voxel CNN~\cite{liu2019point}, where the latter synergizes PointNet with voxel partitions to impose spatial locality.
Despite its premise of sparsity and scalability, this type of structure has not been adopted for large-scale scene-level 3D generation.

\paragraph*{Hierarchy.}
One design choice inherited from 2D diffusion models is if a method employs a multi-stage (typically two-stage) diffusion process to achieve high-resolution generation while under memory and computation constraints. As exemplified in \cite{hui_neural_2022,zheng_locally_2023,zeng_lion_2022,nichol_point-e_2022,karnewar2023holofusion,kim2023neuralfield},
the first stage generates a coarse representation and the second stage refines the output to a higher resolution.
It is noteworthy that such a two-stage approach can use different representations for the lower and the higher resolution, for example~\cite{zheng_locally_2023} uses an occupancy grid in the first stage and an SDF octree in the second stage.

\paragraph*{Types of Supervision.}\label{sec:direct_3D_supervision}
While most of the discussed techniques try to learn 3D shapes or scenes from datasets of 3D shapes or scenes, a different approach is to follow the success of 3D GANs to train a diffusion model directly from datasets of 2D images.
These methods can be considered as a special form of latent diffusion, in which the latent codes capture 3D information, yet the decoder converts the latent codes to 2D observations.
Examples are RenderDiffusion~\cite{anciukevivcius2023renderdiffusion}, Holodiffusion~\cite{karnewar2023holodiffusion}, SSDNeRF~\cite{chen2023single}, GAUDI~\cite{bautista2022gaudi}, Viewset Diffusion~\cite{Szymanowicz2023}, and Diffusion with Forward Models~\cite{forwardmodels}.
In particular, the first four methods adopt an end-to-end training strategy, integrating the learning of the latent codes and the diffusion process into a single training process.
As such, these methods are able to learn the latent codes that are most suitable for the diffusion process, while using the diffusion prior in-the-loop to regularize and improve the convergence of the latent learning.
Notably, such an end-to-end approach has a significant meaning, as it eliminates the need to first train the ground truth ``clean'' 3D representation, which often requires 3D or dense 2D image supervision.
Control3Diff~\cite{gu2023learning} merges diffusion models and GANs to enable training on single-view datasets, such as FFHQ and AFHQ. It leverages EG3D~\cite{chan2022efficient}  to generate numerous pairs of control signals and tri-planes, employing a diffusion model with optional image guidance to learn the prior distributions of tri-planes and camera poses from input images.
In future work, one could leverage more diverse large-scale 2D image datasets, such as LAION, which can be orders of magnitude larger than the largest 3D datasets.

\paragraph*{Conditioning.}
Multiple types of conditioning have been used in the context of 3D diffusion models: text, images, depth maps, sketches, segmentation maps, shape category, or partial or coarse shapes.
%
Most importantly, the task of text conditioning in these models presents unique challenges, primarily due to the scarcity of large-scale 3D datasets with corresponding text descriptions. Existing datasets, such as Text2Shape~\cite{chen2019text2shape}, ShapeGlot~\cite{achlioptas2019shapeglot}, PartIt~\cite{hong2021vlgrammar}, SNARE~\cite{thomason2022language}, and ShapeTalk~\cite{achlioptas2023shapetalk}, are predominantly confined to a limited range of object-level shape categories. This constraint significantly curtails the complexity and diversity of the generative models, thereby limiting their applicability.
To circumvent the dearth of text-shape paired datasets, some studies have resorted to using images as surrogate data. These methods exploit the shared embedding space between text and images to condition the diffusion process. During the training phase, an image representation of the 3D shape is initially obtained through a rendering process. Subsequently, models such as CLIP are employed to derive an image embedding that serves as a conditioning variable for the diffusion model. At test time, a text prompt is processed through CLIP to obtain a text embedding, which can then be used interchangeably with the image embedding. This approach is agnostic to the choice of the underlying generative model. For instance, when initially proposed by Sanghi et al. \cite{sanghi2022clip}, the generative model was based on normalizing flows. Diffusion-based 3D generative methods, such as those presented in\cite{sanghi2023clip,nichol_point-e_2022,zhao2023michelangelo,jun2023shap}, have successfully adopted this strategy.

\paragraph*{Datasets.}
\Cref{tab:geometry_gen_summary} also includes the specific datasets that were used by the different papers. We do not provide an in-depth discussion of these datasets but refer the reader to Sec.~\ref{sec:data} that reviews the most important datasets used by diffusion techniques overall.

\paragraph*{Summary.}
The complexity of the design space in ``Direct 3D Generation via Diffusion Models'' stems from the lack of standardized data representations and architectures for 3D content.
The 3D data landscape is characterized by its problematic sparsity coupled with a large spatial span, making it a challenging candidate for any single, standardized architecture capable of effective spatial reasoning.
Consequently, researchers have ventured into a multitude of design choices, ranging from the type of output and 3D shape representation to the architecture of the diffusion model and supervision strategy.
In this section, we have attempted to identify the most fundamental design choices explored by recent publications, shedding a light on their impact and the tradeoffs involved.
Despite these advancements, it is imperative to acknowledge that 3D assets remain substantially less abundant compared to their 2D counterparts.
While future work will be able to build on larger datasets, such as Objaverse~\cite{objaverse} and Objaverse-XL~\cite{ObjaverseXL}, there is still a large discrepancy in dataset size between 2D and 3D diffusion models.
This discrepancy underscores the need for innovative methodologies that can bridge this gap. In the ensuing section, we will explore how 2D diffusion models can be leveraged to facilitate the generation of 3D content.

\begin{figure}[t]
	\begin{center}
        \includegraphics[width=\linewidth] {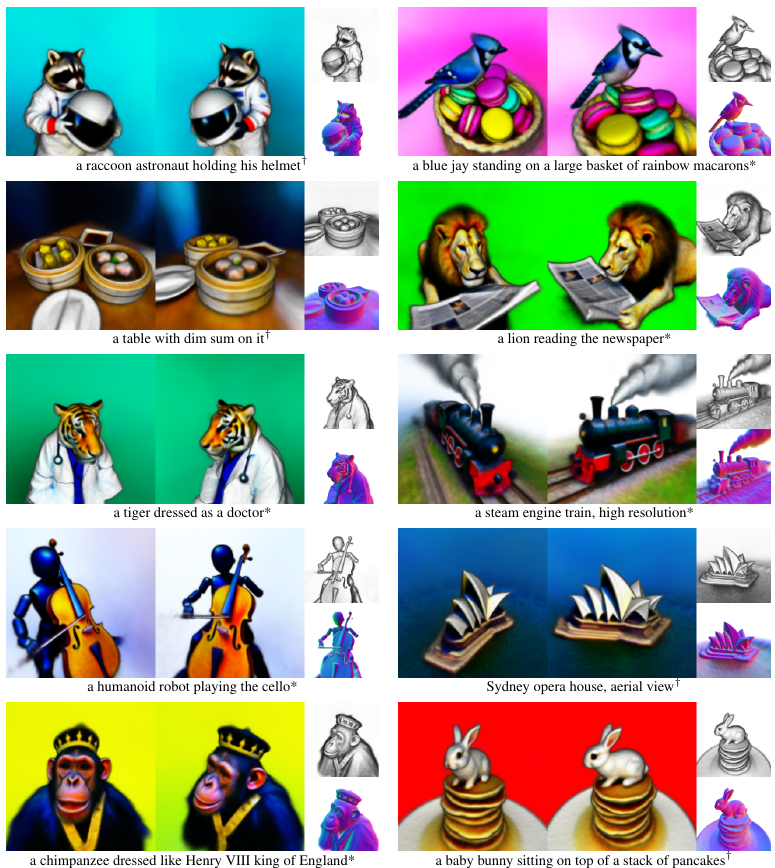}
	\end{center}
	\caption{\textbf{Text-to-3D Generation.} Example text-to-3D generations from DreamFusion~\cite{poole2022dreamfusion} with corresponding input text-prompts. Each synthesized 3D model is rendered in two views with untextured renders and normals shown to the right. Image adapted from~\cite{poole2022dreamfusion}.}
	\label{fig:dreamfusion}
\end{figure}

\subsection{Leveraging 2D Diffusion Models for 3D Generation}

The exploration of 2D diffusion models for 3D generation stems from the notable advancements in image synthesis conditioned on text. These strides owe their success to abundant text-aligned image datasets and scalable model architectures. While similar techniques have been attempted for 3D generation (as described in the previous section), the scarcity of 3D data and the lack of well-explored denoising architectures present significant challenges. Notably, large-scale datasets such as LAION-5B~\cite{schuhmann2022laion5b}, containing 5 billion text-image pairs, dwarf the largest available text-3D dataset, Objaverse-XL~\cite{ObjaverseXL}, with its mere 10 million paired samples.

To surmount the limitations imposed by limited 3D data and architectures, innovative strategies leveraging 2D image priors for 3D generation have emerged. These techniques distill multi-view geometry understanding, either implicitly learned from large-scale image diffusion models or explicitly via image diffusion models additionally conditioned on input camera poses and trained or finetuned on multi-view datasets.

\subsubsection{Methods}
\paragraph*{Text-to-3D using Pre-trained Image Diffusion Models.}
Utilizing pre-trained image diffusion priors, DreamFusion~\cite{poole2022dreamfusion} stands as a prime example, achieving groundbreaking text-to-3D generation. This innovative approach harnesses established image priors, allowing for zero-shot synthesis of intricate 3D objects, as vividly demonstrated in Fig.~\ref{fig:dreamfusion}. The method revolves around optimizing a 3D model represented as a Neural Radiance Field (NeRF) through a specialized image space loss. This loss function, tailored to leverage pre-trained image diffusion models as priors, assigns lower values to plausible images, ensuring the coherence of generated 3D objects. This process of sampling through gradient descent optimization is known as Score Distillation Sampling (SDS).
Formally, a 3D scene representation parameterized by $\phi$ is rendered by a differentiable generator $g$ at a sampled camera pose, generating an image $g(\phi)$. Gaussian noise $\eta$ is then injected into the image and passed through an image diffusion prior parameterized by $\theta$, with text condition \(\varcond\). 3D scene parameters are then updated via the gradient:
\begin{equation}
    \nabla_\phi \mathcal{L}_\textrm{SDS} (\theta, g(\phi)) = \mathbb{E}_{t,\epsilon} \Bigl[ w(t)(\hat{\epsilon}_\theta(\diffusevar_t; \varcond, t) - \eta) \frac{\partial g(\phi)}{\partial \phi}\Bigr].
\end{equation}
Importantly, this technique has broader applicability, extending beyond 3D synthesis as explored by Hertz et al.\cite{hertz2023delta} for image editing.
Crucially, this optimization through 3D model and differentiable rendering guarantees that each optimized image adheres to multi-view constraints, ensuring a coherent 3D model. This methodology, embraced by subsequent works such as Score Jacobian Chaining~\cite{sjc}, Magic3D~\cite{lin2023magic3d}, and Latent-NeRF~\cite{metzer2023latent}, maintains a similar philosophy while exploring diverse underlying representations for the optimized 3D model.

However, direct SDS application for 3D generation poses challenges, notably the \textit{Janus problem}, where radially asymmetric objects exhibit unintended symmetries, like multiple
faces on a human head.
To mitigate this artifact, researchers in works like DreamFusion~\cite{poole2022dreamfusion}, Score Jacobian Chaining~\cite{sjc}, Latent-NeRF~\cite{metzer2023latent}, and Magic3D~\cite{lin2023magic3d} employ strategies like view-dependent prompting, e.g., \emph{``top-view of ...''}, and additional regularization terms. Alternatively, learning 3D representations from multi-view diffusion models ensures consistency across diverse views.

Additionally, SDS-based 3D generation demands operation at unusually high guidance scales, influenced by the mode-seeking behavior of the loss function. Operating beyond the typical range of the pre-trained image prior leads to saturation artifacts and limited diversity in outputs. ProlificDreamer~\cite{wang2023prolificdreamer} tackles these challenges through Variational Score Distillation, a generalized SDS version. This method incorporates finetuning of the image model using LoRA~\cite{hu2021lora} during per-scene optimization, with the intention of mitigating both the Janus problem and the need for high guidance scales.

\paragraph*{Adapting Image Models for Multi-view Synthesis.} The efficacy of image diffusion models for 3D generation, as exemplified by DreamFusion~\cite{poole2022dreamfusion}, has prompted subsequent investigations. Follow-up studies have underscored the enhanced generation quality achievable through the finetuning of pre-trained image priors with 3D data.
Zero-1-to-3~\cite{Liu2023Zero1to3ZO} finetunes a pre-trained text-to-image model to add camera pose conditioning. This finetuning is done on pairs of multi-view images rendered from the Objaverse dataset~\cite{objaverse}. The resulting model takes as input an image and the camera parameters of a new viewpoint, and returns an image rendered at the novel view. This method has also been proven to work at scale, as Deitke \etal~\cite{ObjaverseXL} train Zero-1-to-3-XL in a similar fashion using the larger Objaverse-XL dataset. This flavor of pose-conditioned image diffusion model can also be trained from scratch on multi-view image data, as shown by 3DiM~\cite{3dim}. Performing DreamFusion-like 3D aggregation using a diffusion model with these added pose conditioning signals helps reduce the need for certain tricks like view-dependent text prompting. Unfortunately, these single-view models still suffer from many of the same artifacts as standard DreamFusion due to the fact that only a single image is queried a time (and, therefore, multiple queried viewpoints may propagate contradictory signals into the 3D model).
Newer methods aim to resolve this: MVDream~\cite{mvdream} finetunes a pre-trained image diffusion model to create a multi-view diffusion model, capable of generating a \emph{set} of geometrically consistent images of the same object at four fixed camera poses, from an input text prompt. This is achieved through the addition of a 3D self-attention module trained on multi-view images rendered from the Objaverse~\cite{objaverse} 3D dataset. The resulting multi-view diffusion model can be directly used for 3D generation through SDS. Since the multi-view diffusion model outputs images from four orthogonal azimuth angles, rather than from a single view at a time, this method provides a principled remedy to the aforementioned Janus problem. SyncDreamer~\cite{syncdreamer} approaches multi-view diffusion by grounding features from each generated view into an explicit 3D feature space. Using a 3D-aware attention mechanism, SyncDreamer synchronizes the intermediate states between the diffusion paths across different viewpoints, establishing 3D correspondence between them.

\paragraph*{3D-Aware Image Diffusion.} Although finetuning higher-dimensional models on top of large pre-trained T2I models helps simplify the training process, the choice to build off a pre-trained model can limit control over the architectural design space. Another line of work explores training models from scratch, such that architectures may be 3D-aware or use more explicit multi-view reasoning to provide conditioning signals. GeNVS~\cite{genvs} tackles single-image novel-view synthesis by training a 3D-aware conditional diffusion model that incorporates geometric priors in the form of a 3D feature volume obtained from an input image (or images, when performing autoregressive generation). A feature image rendered from the feature volume at the novel viewpoint provides 3D-aware conditioning to the diffusion model. NerfDiff~\cite{nerfdiff} takes a similar approach, providing  PixelNeRF~\cite{pixelnerf} renderings at novel views as conditioning to a 3D-aware diffusion model. SparseFusion~\cite{sparsefusion} and Sparse3D~\cite{sparse3d} use an epipolar feature transformer~\cite{epipolarfeaturetransformer} to provide a conditioning signal to a finetuned image diffusion model.

In a separate vein, 3D-aware image generation has also been explored through the perpetual view generation problem introduced by InfiniteNature~\cite{infinite_nature_2020, li2022_infinite_nature_zero}. For example, given only a text prompt, SceneScape~\cite{Fridman2023SceneScapeTC} generates 3D-consistent videos depicting static scenes rendered from a specified camera trajectory. To achieve this, they leverage a pre-trained T2I model, and construct a unified mesh representation of the scene, along with the video generation process. DiffDreamer~\cite{cai2023diffdreamer} trains a conditional diffusion model that takes a single input image and generates renderings of a specified camera trajectory flying into the scene. Although these tasks seem similar, flying into a scene is more challenging because this problem not only requires 3D-multi-view consistent out-painting but also super-resolution, as novel details of a previously seen structure become visible in later frames.

\subsubsection{Applications} \label{ssec:3D_Applications}
\paragraph*{3D Editing.}
Following the success of InstructPix2Pix~\cite{brooks2023instructpix2pix} for instruction-based image editing, InstructNeRF2NeRF~\cite{haque2023instruct} achieves similar results for editing 3D scenes. Similar to SDS, the diffusion model (in this case, a text-and-image conditioned model that edits images) is queried repeatedly through optimization, and the model outputs are propagated back into a the NeRF scene. A few modifications are made to the standard SDS formulation, however. First, the diffusion model is queried not once, but rather multiple times, such that the fractionally noised image can be sampled to a clean output image (as in~\cite{sparsefusion}). Then, to derive a loss, instead of directly comparing the noise estimate to the injected noise, the clean image is compared to the original scene rendering. Finally, this loss is not applied for one image at a time (as in DreamFusion), but rather at a randomly shuffled set of rays from the set of captured views (which is standard in NeRF optimization). This effectively amounts to iteratively updating the dataset of images used for training the NeRF, and is similar in spirit to the method propose in SNeRF~\cite{SNeRF} for 3D style transfer.

\begin{figure}[t]
	\begin{center}
        \includegraphics[width=\linewidth] {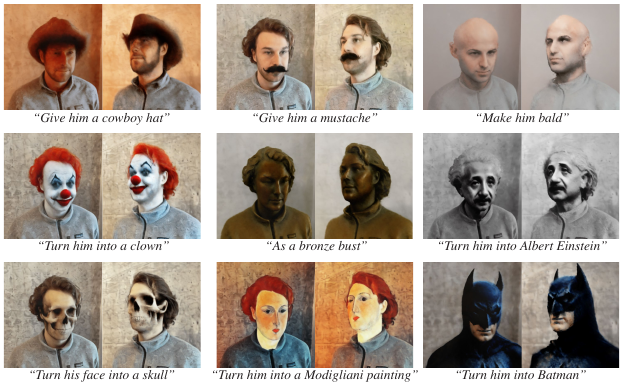}
	\end{center}
	\caption{\textbf{Instruction-based 3D Editing}. Example instruction-based 3D edits from InstructNeRF2NeRF~\cite{haque2023instruct} with corresponding text-prompts. All edits are performed over the same input reconstructed NeRF scene. Image reproduced from ~\cite{haque2023instruct}}
	\label{fig:instructn2n}
\end{figure}

Fig.~\ref{fig:instructn2n} shows example edits of the same initial 3D scene with the corresponding text-based instructions. InstructNeRF2NeRF can handle a diverse set of instructions while performing multi-view consistent 3D edits. While InstructNeRF2NeRF enables diverse holistic and contextual edits with high fidelity, it can often be bottlenecked by the performance of InstructPix2Pix: If a certain edit is not possible in 2D or is too inconsistent, it will not be reflected in 3D. Similarly, one occasional failure mode of InstructPix2Pix is over-editing, i.e., when unintended parts of the scene are modified. To resolve this, and have guarantees on localized editing, one may inherit insights from methods like DreamEditor~\cite{zhuang2023dreameditor} and Vox-E~\cite{sella2023voxe} that can detect the region to be edited using the underlying attention maps from the diffusion model.


\paragraph*{Scene Generation.}
Locally conditioned diffusion (LCD)~\cite{Po2023Compositional3S} enables controllable 3D generation with intuitive user inputs. LCD transforms user-defined bounding boxes with corresponding text prompts and generates 3D scenes matching the desired layout. This is achieved using a modified SDS-based pipeline that takes the user input as conditioning. The method allows explicit control over the size and position of individual scene components while ensuring seamless transitions between them. Text2Room~\cite{hollein2023text2room} generates textured 3D meshes of indoor rooms from a given text prompt. Although it also leverages a pre-trained image prior, 3D synthesis is not enabled through SDS. Instead, a pre-trained monocular depth estimator is used to provide 3D information. The method generates a textured 3D mesh by iteratively inpainting the scene at randomly sampled camera angles. During each inpainting step, a monocular depth estimator is used to create a 3D mesh of the new scene content, which is then merged with the rest of the existing geometry.

%

\section{Towards 4D Spatio-temporal Diffusion}\label{sec:4D_generation} 

The capabilities of video generation and editing tools are rapidly advancing (see Sec.~\ref{sec:video}), but the underlying T2V models often lack a mechanism to model long-range temporal consistency or consistent 3D structure of their objects and scenes. 3D generative tools (see Sec.~\ref{sec:spatial}) are specifically designed to enable spatial reasoning. Yet, these tools are limited to modeling static scenes with rigid (camera) motion. We consider video generation and editing that have an understanding of the underlying 3D structure of a (dynamic) scene as 4D in the sense that they combine 3D structure and time, including non-rigid deformation and general motion. In this realm, we discuss articulated object and avatar generation, motion generation, and more general 4D generation and editing tools in the following.

\subsection{Articulated Avatars, Animals, and Objects}
\label{sec:4D:articulation}

Articulation refers to a class of non-rigid motions in which the deformation of an object is composed of locally rigid transformations. 
Many objects in everyday life are articulated, including human faces and bodies as well as animals.  
Articulated objects can be intuitively controlled in a manual manner or using motion capture, for example using skeleton-based or template-mesh-based handles. This set of techniques has been popular and broadly applied in the field of computer graphics for the last few decades. 

Generating articulable 3D avatars from a given text prompt is a popular line of research in this area. This class of methods often generates 3D digital humans by constraining the outputs to follow a given parameterized model. For example, DreamHuman~\cite{dreamhuman} generates articulable human avatars using imGHUM~\cite{imghum}, an implicit statistical 3D human pose and shape model. Bergman et al.~\cite{bergman2023articulated} synthesize articulated 3D heads by optimizing the geometry and texture of a 3D morphoable model and TADA!~\cite{tada} optimizes over a human model derived from SMPL-X~\cite{SMPL-X:2019}. 
DreamFace \cite{Zhang2023} is a CLIP-based selection approach for animatable head avatar generation from text prompts. It first chooses coarse proxy geometry and then refines it for consistency with the text prompt using SDS loss and adds the hair. 
%
TECA \cite{zhang2023teca} synthesises 3D human avatars composed of mesh-based head (generated first) and other elements based on NeRF (hair, garments and accessories; added afterwards using SDS loss and guidance from segmentation) from text descriptions; the texture is inpainted with diffusion models. Animations of the generated avatars can be performed leveraging a human parametric model \cite{SMPL-X:2019}. 
DreamAvatar~\cite{cao2023dreamavatar} and AvatarCraft~\cite{jiang2023avatarcraft} also focus on the creation of articulable avatars from text prompts and support body articulation.
4D Facial Diffusion Model \cite{Zou2023arXiv} generates facial expressions of a template face mesh.
The AvatarStudio method~\cite{Mendiratta2023} creates text-guided stylizations of high-quality 3D neural face avatars using a new view- and time-consistent score distillation sampling. 

Diffusion-based generation of more general 3D articulated objects has been explored by NAP \cite{Lei2023}. This system uses a tree parameterization for modeling the irregular distribution of articulated objects across different datasets and supports unconditional and conditional articulated shape generation (e.g., conditioned on object parts or joints). 
ARTIC3D \cite{Yao2023artic3d} leverages a generative 2D diffusion model prior to learning articulated and animatable 3D animal shapes. 
Given a sparse and (not curated and without 2D or 3D annotations) image collection containing 10--30 images of an animal species, ARTIC3D estimates camera viewpoints, pose articulations, part shapes and texture for each observed instance; no shape initializations, pre-defined 3D skeletons or shape templates are required, in contrast to previous works in this area \cite{cmrKanazawa18,Sidhu2020, Yang2021, yao2022lassie,wu2023magicpony}.

\subsection{3D Human Motion Generation}
In recent years, deep learning has made remarkable strides in advancing character animation. Thanks to the availability of large human motion datasets to the public, researchers in character animation have been training data-driven models to predict character motion, conditioned on the motion history, agent observations or various forms of user commands. However, a persistent challenge in developing predictive motion models is that the mapping between input conditions and output motions is rarely one-to-one. For example, when a character is instructed to approach a chair, they have multiple options, such as walking around it, moving it, or sitting on it. This challenge necessitates the adoption of generative models. Indeed, generative models provide a principled solution to learn a conditional data distribution, making them well-suited for modeling many-to-many mappings inherent in human motion synthesis.

A significant body of work has explored approaches based on Generative Adversarial Networks (GANs) and Variational Autoencoders (VAEs) to synthesize motion trajectories \cite{barsoum2018hp} or train auto-regressive motion priors \cite{ling2020character,rempe2021humor}. While these methods have shown varying degrees of success, the full benefits of generative models were not always realized due to issues like mode collapse, posterior collapse or training instability. However, in early 2022, diffusion-based approaches entered the field of character animation and rapidly became the preferred choice among generative models. Unlike Conditional VAEs (CVAE) and GANs, diffusion models excel at modeling multimodal distributions and offer ease of training. Additionally, diffusion models allow for flexible motion editing during inference using techniques such as ``In-painting'', which aligns well with the needs of digital content creation across a wide range of computer graphics applications.

Nevertheless, the successful application of diffusion in motion synthesis heavily relies on the quality and quantity of training datasets. Furthermore, for conditioned models, a demanding prerequisite is the existence of paired training datasets that establish a clear correspondence between the conditioning information and 3D human motion. These constraints can indeed limit the applicability of diffusion models. Currently, various human motion datasets exist, coupled with diverse conditions, including text \cite{Guo_2022_CVPR,guo2020action2motion}, music \cite{li2021learn}, audio \cite{ferstl2018investigating}, video \cite{h36m_pami,mono-3dhp2017}, scene descriptions \cite{araujo2023circle,guzov2021human,zheng2022gimo,yuan2019ego,Zhang:ECCV:2022} and objects \cite{GRAB:2020,fan2023arctic,bhatnagar22behave,li2023omomo}. Continuing to develop and share such large-scale, diverse, and high-quality paired motion datasets is a valuable investment in the relevant research and development communities.

Different motion representations have been used in prior work without a clear advantage of one over the other. While all assume an underlying body model, some use a minimal representation such as the 6D joint rotations \cite{tseng2023edge,kim2023flame}, while others allow a redundant representation by having both positions and rotations \cite{Tevet2023,chen2023executing}. It is also a common practice to include contact information as part of the motion representation to address concerns like foot sliding or surface penetration \cite{tseng2023edge}. Some studies have suggested employing redundant representations and using self-consistency to improve learning of diffusion models \cite{Tevet2023}. Furthermore, learning an encoded motion representation has also shown benefit in training diffusion models \cite{chen2023executing, jiang2023motiongpt,ao2023gesturediffuclip}.

Despite the initial success of diffusion-based approaches in the early exploration phase, several recurring issues have surfaced in motion synthesis applications. Notably, diffusion models, like most machine learning techniques, lack an inherent ability to precisely satisfy constraints, often resulting in motion artifacts such as violations of physical and geometric properties, or failure to follow the control signals or specifications provided as input. Several proposed methods aim to address this concern, including gradient guidance \cite{rempe2023trace, karunratanakul2023gmd}, learning of conditions via supervision \cite{li2023omomo}, direct editing during inference \cite{Tevet2023}, or refining the model through reinforcement learning processes \cite{han2023amd}. While these methods have shown varying degrees of success in controlling output motion, they cannot guarantee precise constraint satisfaction and controllability. 

The remaining part of the subsection highlights a few active research directions in motion synthesis and the representative papers that advanced the areas.

\begin{figure}[t] 
	\begin{center} 
        \includegraphics[width=\linewidth] {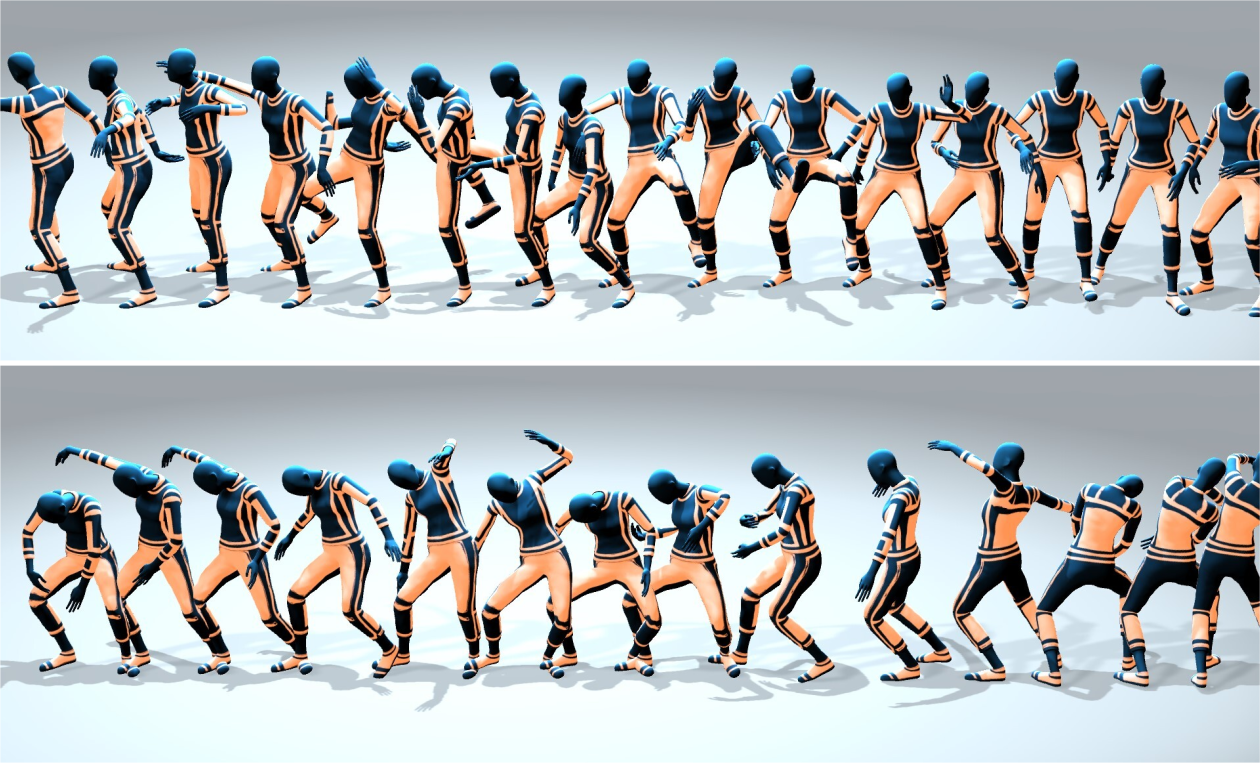} 
	\end{center} 
	\caption{\textbf{Motion Generation.} Dances synthesised from audios by the approach of Alexanderson {et al.}~\cite{Alexanderson2023} for Locking (top row) and Krumping (bottom row) dance styles. Image reproduced from \cite{Alexanderson2023}.} 
	\label{fig:Alexanderson_etal_2023} 
\end{figure}

\paragraph*{Motion Generation Conditioned on Time-series.}
A number of recent works focus on generating motion synthesis from a time-series input, such as audio, music, or text. EDGE \cite{tseng2023edge} is a transformer-based diffusion model paired with a pre-trained music feature extractor Jukebox \cite{Dhariwal2020}---acting as cross-attention context---for realistic dance motion generation. EDGE offers versatile editing functionality and arbitrary long sequences. This is possible by replacing the known regions with forward-diffused samples of the provided constraint (through masking), which is an increasingly popular technique enabling editability at test time without the need for model retraining. Listen, Denoise, Action!~\cite{Alexanderson2023} is a method based on a stack of Conformers \cite{Zhang2022} for dance motion generation from audio signals (see Fig.~\ref{fig:Alexanderson_etal_2023}). It takes acoustic feature vectors and an optional style vector as inputs. The method enables control over motion style with classifier-free guidance and dance style interpolation thanks to the product-of-experts ensembles of diffusion models (guided interpolation of different probability distributions). MoFusion \cite{dabral2022mofusion} conditions human dance generation on Mel spectrograms. The advantage of this method is that the context embedding layer of the MoFusion architecture learns a suitable injection of the audio signal to the feature space of a U-Net, in contrast to MFCC features, for example, which offer less flexibility. 

\paragraph*{Motion Generation with Spatial Constraints.}
Integration of spatial constraints into 3D human motion generation with diffusion models is another active research direction. GMD \cite{karunratanakul2023gmd}, for example, offers two elaborate mechanisms to allow spatial conditioning including trajectory, keypoints and obstacle avoidance objectives. NIFTY's architecture \cite{kulkarni2023nifty} includes an object interaction field that guides the motion generator to enable the support of human-object interactions. AGRoL~\cite{Du2023} is an MLP-based architecture with a conditional diffusion model for full-body motion synthesis accepting a sparse upper-body tracking signal. The method supports real-time inference rates, which places them among only a few techniques well-suitable for AR and VR applications. InterGen \cite{Liang2023} is a model generating interactive human motions. The paper considers the case of two virtual characters constraining each other's movements.  
InterGen's objective is reflected in its architecture: It contains cooperative denoisers that share weights and a mutual attention mechanism, which improves the motion generation quality. Additional interaction regularizers (losses) with a damping schedule allow modeling of complex spatial relations between the generated motions. Similarly, Shafir et al.~\cite{Shafir2023} train a slim communication layer to synchronize two pre-trained motion models using only a few training examples. PhysDiff \cite{Yuan2023} goes a different way beyond geometric priors, namely physics-guided motion diffusion: It adds a physics-based motion projection step into the diffusion process of existing methods that mitigates such artefacts as floating in the air, foot-floor penetration and foot sliding. The projection step of PhysDiff is a human motion imitation policy controlling a virtual character in a physical simulator and enforcing the physical constraints.

\paragraph*{Synthesis of Long Motion Trajectories.}
Being able to generate arbitrarily long motion sequences is crucial to many online applications. The TEDi approach \cite{zhang2023tedi} entangles the time-axis of diffusion and temporal motion axis: In each diffusion step of their stationary motion buffer (aka~pre-defined motion generation window), they remove a clean frame at the beginning of the buffer and append a new noise vector to its end. A different approach is pursued by DoubleTake, \cite{Shafir2023} that extends MDM for long-term motion synthesis without additional training. Their sequential composition approach concatenates short motion sequences with a diffusion blending step (performed in a zero-shot manner): In the first ``take'', their method generates motion batches and the second ``take'' refines the transitions between individual motions. Their motion prior is trained from short 3D motion clips only, and the method allows individual control of each motion interval. MoFusion \cite{dabral2022mofusion} can generate long motion using seed conditioning in a sliding window fashion. The provided seed motion frames are first corrupted by noise, while the remaining ones that need to be generated are initialised with random noise. At each denoising step, the seed frames are masked out, which eventually leads to generated motions complementing the seed motions. A similar policy is also applicable to MotionDiffuse \cite{zhang2022motiondiffuse}, MDM \cite{Tevet2023} and EDGE \cite{tseng2023edge}. In Alexanderson {et al.}~\cite{Alexanderson2023}, translation-invariant policy to encode positional information enables better generalisation to long sequences. 

\paragraph*{Motion Generation Based on Multiple Modalities.}
Many motion generation algorithms leverage inputs with multiple modalities. For example, gesture generation approaches use speech as a condition along with text for style; the attention mechanism is leveraged to synchronise the gestures to the speech. The approach of Deichler {et al.}~\cite{Deichler2023arXiv} proposes a contrastive speech and motion pre-training module that learns joint embedding of speech and gesture; it learns a semantic coupling between these modalities. DiffuseStyleGesture \cite{yang2023diffusestylegesture} is an audio-driven co-gesture generation approach that synthesises gestures matching the music rhythm and text descriptions based on cross-local and self-attention mechanisms. It uses classifier-free guidance to manipulate the initial gestures and interpolate or extrapolate them. EMoG \cite{Yin2023} decomposes the generation problem into two steps, i.e., joint correlation and temporal dynamics, and shows that explicitly predicting joint correlation improves generation quality. 
%
%
DiffMotion \cite{Zhang2023DiffMotion} is a two-stage framework for speech-driven gesture synthesis. 
Its auto-regressive temporal encoder (LSTM-based) conditions the diffusion module on temporal dynamics extracted from the preceding human poses (gestures) and acoustic speech features.

\subsection{4D Scene Generation and Editing}
\label{sec:4D:general}

\paragraph*{4D Scene Generation.} 
General 4D scene generation implies that no strong priors are used about types of objects and possible non-rigid movements, or weak priors are combined with stronger ones (e.g., in a compositional setting). We discuss two approaches in this category that address this most general setting among all, i.e., MAV3D and Learnable Game Engines (LGE). 

MAV3D~\cite{Singer2023TextTo4DDS} (see Fig.~\ref{fig:teaser}, top-right) extends DreamFusion \cite{poole2022dreamfusion} with the time dimension for non-rigid NeRF scenes generated from text. 
Similar to the 3D case, it is difficult to acquire large datasets of 4D scenes that could be used to train non-rigid scene generators, especially with paired textual annotations. 
Hence, MAV3D relies on a pre-trained text-to-video diffusion model \cite{singer2022make} (see Sec.~\ref{sec:video}) as a non-rigid ``world scene prior'' that provides learnt distributions of multi-view scene projections (this is in contrast to \cite{poole2022dreamfusion} using a text-to-image model). 
LGE \cite{Menapace2023} is a neural model with game-engine-like features learned from annotated videos. 
It follows a scene composition approach with the elements of playability (in the sense of Playable Environments \cite{Menapace2022PlayableEnvironments}), scene editability and learning game engine functionality from data. 
LGE supports the generation of compositional 4D NeRFs (e.g., of a tennis match or a Minecraft-like game) conditioned on a single RGB image and a wide spectrum of conditioning temporal signals (such as learned human actions, object locations, and textual action descriptions). 
In LGE, the diffusion-based animation module predicts scene states conditioned on user-provided actions, and the synthesis module renders them from desired viewpoints. 
Compared to MAV3D, LGE generates scenes of higher visual fidelity and with finer-grained control, at the cost of scenario-specific datasets with textual annotations. 

\paragraph*{4D Editing.} 
Control4D \cite{Shao2023Control4DDP} can edit 4D human portraits from text inputs provided as implicit Tensor4D \cite{Shao2023tensor4d} scene representations. 
The method utilizes a 2D diffusion-based prior (editor) to train a 4D GAN that is applied at test time to the rendered 2D views of the dynamic (animated) portrait scenes. 
Note that the diffusion-based editor does not directly operate on the rendered images, and this design choice is done for the 4D GAN to ensure that the edits are temporally consistent. 
While the results of Control4D are also photo-realistic, it can be noticed that they preserve only a few characteristics of the original identity, i.e., the edits are dominated by the textual prompts, especially in the head area. 
A critical reader might ask to what extent the input portraits influence the edited results and whether they could not have been obtained by further refining the textual prompts. 

This aspect is substantially improved in another work titled AvatarStudio \cite{Mendiratta2023} that performs editing of human head avatars in a photo-realistic and temporally consistent manner while preserving the initial identity (see Fig.~\ref{fig:teaser}, second row on the right). 
AvatarStudio accepts as input a high-resolution $360^\circ$ non-rigid NeRF and uses an LDM \cite{rombach2022high} fine-tuned with viewpoints and time stamps randomly sampled from the NeRF volume as well as view-and-time aware SDS loss with a model-based guidance. 
The method edits the full head performance in the canonical space and propagates the edits to all time steps via a pre-trained deformation network. 
AvatarStudio enables personalized and photo-realistic edits that preserve the initial head identity while transferring (or mixing in) the visual appearance features associated with the prompted identity (e.g., ``Vincent van Gogh'' or the literature/movie character ``Cruella de Vil'') or qualitative scene descriptions (e.g., ``bronze bust'', ``marble statue'' or ``blue hair''). 


\section{Data}\label{sec:data}


In this section, we briefly review datasets commonly used for training and evaluating 3D and 4D diffusion models.

\paragraph*{Image Datasets.}
%
Image datasets play a pivotal role in the training and validation of diffusion models for visual computing applications.
High-quality, diverse, and large-scale image datasets, accompanied by rich semantic information, such as class or semantic (instance) labels \cite{deng2009imagenet,lin2014microsoft}, ensure that trained models can generalize across many image modalities.
As the demand for labeled images grows, the scale of such datasets continues to expand.
In particular, the combination of image and text data has been an important source of training data in the context of diffusion models, since such data can be crawled from the web at large scale and typically does not rely on manual annotations.
Notable large-scale image datasets include the Wikipedia-based Image Text (WIT)~\cite{Srinivasan2021WITWI}, a curated set of 37.6 million rich image--text examples with 11.5 million unique images across 108 Wikipedia languages, the LAION-400M open dataset~\cite{laion400m} of 400 million CLIP-Filtered image-text pairs, and its successor LAION-5B~\cite{schuhmann2022laion5b} which expands the collection to 5.85 billion.




\paragraph*{Video Datasets.}
Publicly available video datasets with textual descriptions, such as WebVid-10M~\cite{Bain21} and HD-VILA-100M~\cite{xue2022hdvila}, are of great importance for training video diffusion models. 
These datasets contain 10M and 100M text--video pairs, respectively, which is more than an order of magnitude smaller than available text--image datasets. 
For this reason, many T2V models are trained with both images and videos, treating images as individual video frames. 
Alternatively, a pre-trained T2I model can be refined with a smaller amount of training data (see Sec.~\ref{sec:video} for more details). 
At the same time, these datasets can be augmented with the vast amount of unlabeled video data available on the web.

\paragraph*{Shape Datasets.}
In contrast to their image-based counterparts, 3D datasets are often constrained by the paucity of training samples, primarily due to the high cost associated with obtaining 3D models.
A key dataset in this domain is ShapeNet~\cite{chang2015shapenet}, which comprises 51.3k models across 55 object categories and has been enriched with part segmentation and textual annotations~\cite{achlioptas2019shapeglot,chen2019text2shape,koo2022partglot,achlioptas2023shapetalk}.
Another widely-used dataset is Amazon Berkley Objects (ABO)\cite{collins2022abo}, which emphasizes texture quality and encompasses 8k 3D models from 63 classes, serving as a training ground for object-level geometry and appearance generative models. 
While these datasets offer synthetic models that provide abundant training data and ground-truth geometry, they also introduce a synthetic-to-real gap.
To mitigate this, CO3D~\cite{reizenstein2021common} and OmniObject3D~\cite{wu2023omniobject3d} acquired large-scale real objects. 
While CO3D contains 19k objects, it only provides multi-view captures, whereas OmniObject3D offers 6,000 high-quality scanned 3D objects across 190 categories, complete with textured mesh and point clouds.
However, the most ambitious endeavor in this space of object-level 3D data is Objaverse and Objaverse-XL~\cite{objaverse,ObjaverseXL}, boasting over 10M 3D models sourced from Sketchfab~\cite{sketchfab}, accompanied by a large-scale text corpus.
Despite its unprecedented scale, the dataset presents challenges due to the heterogeneous quality of 3D models and text descriptions, as well as non-uniform data distribution, thus complicating the effective utilization of these large-scale resources.
For a scene-scale generation, data collection is even more challenging.
While seminal datasets such as ScanNet~\cite{dai2017scannet}, Matterport3D~\cite{Matterport3D}, ScanNet++~\cite{yeshwanth2023scannet++}, and RealEstate10k~\cite{StereoMagnification} rely on 3D scanning real environments, others focus on building on crowd-sourced web designs for scene modeling, such as 3D Front~\cite{fu20213d}.
In practice, however, many existing approaches in this space do not (yet) generate an actual 3D scene but rather a video rendering of the scene, which requires the camera trajectory and the corresponding RGB(D) images for supervision.
VizDoom~\cite{kempka2016vizdoom}, Replica~\cite{straub2019replica}, Carla~\cite{dosovitskiy2017carla}, VLN-CE~\cite{krantz2020beyond}, and ARKitScenes~\cite{baruch2021arkitscenes} are some examples of such datasets, where the training sequence is generated through a simulated or captured camera motion in synthetic or reconstructed static 3D scenes.
However, these datasets are still severely limited in diversity and annotations.
The autonomous driving industry is in an ideal position to capture diverse real-world data; however, such data is typically treated as a proprietary asset and not publicly available.
We also provide references to smaller datasets that have been used in approaches mentioned in Sec.~\ref{sec:direct-3d-generation}:
BuildingNet~\cite{selvaraju2021buildingnet}, Pix3D~\cite{sun2018pix3d}, Photoshape~\cite{park2018photoshape}, CLEVR~\cite{johnson2017clevr}.
These datasets complement ShapeNet in terms of geometric scale and complexity, as well as rendering realism, and annotation availability.











\paragraph*{4D and Human Motion Datasets.}


%
%



Another important set of datasets are those focusing on 3D capture in motion \cite{li20214dcomplete,bozic2020deepdeform}. 
Although the human body spans many possible motions, 3D human motion is expensive to acquire, and hence datasets remain scarce.
In the context of humans, most current sizable datasets are based on AMASS \cite{mahmood2019amass}, which is a collection of other datasets acquired by different technologies, all aligned to a uniform representation \cite{SMPL:2015}. 
In total, these recordings contain over $40$ hours and comprise 11k sequences; however, this alignment does not include any labeling. 
Hence, other datasets \cite{plappert2016kit,BABEL:CVPR:2021,Guo_2022_CVPR} have sought  to textually annotate it. 
These datasets include richer and cleaner data, with more elaborate textual descriptions. 
Although already large, these datasets still do not capture the full expressiveness of human motion.
%
A more elaborate dataset has recently been released \cite{lin2023motionx}, consisting of an order of magnitude more data, and more expressive annotations that include hand motions and facial expressions. As literature exploring the capabilities of this dataset is still in its infancy, it is too soon to predict the impact it will have on the field.
At the same time, higher-fidelity animation sequences from 3D scans or multi-view capture setups have been released in the context of human faces~\cite{giebenhain2022nphm,kirschstein2023nersemble} and bodies~\cite{peng2021neural,isik2023humanrf,liu2021neural,habermann2021,deepcap,Xu_monoperfcap}.
Although these datasets are of high quality for each captured instance, the costly recording process limits the largest of such datasets to several hundred captured people. 
%
%
A new dataset with audio and high-quality 3D motion capture with various dance genres for 3D human motion generation tasks was introduced in Alexanderson {et al.}~\cite{Alexanderson2023}. 
AIST++ \cite{li2021learn} is another dataset of people dancing widely used in 3D human motion generation. 

The CIRCLE dataset \cite{araujo2023circle} utilizes optical motion capture and virtual reality to capture 3D human motions in cluttered indoor environments, paired with an egocentric view of the scene.
A very large dataset of speech-aligned 3D face, body and hand motion extracted from talk show host videos is presented in~\cite{3dconvgesture_2021}. 
Finally, the OMOMO dataset \cite{lu2023large} contains more than 10 hours of full-body manipulation of various objects, including pushing furniture, carrying household objects, manipulating cleaning equipment and more.



\section{Metrics}\label{sec:metrics}



This section briefly discusses the metrics used for evaluating the reviewed methods with diffusion models. 

\paragraph*{Image Quality and Diversity.} Robust evaluation of image diffusion models requires the use of specific metrics that can effectively capture both the quality and diversity of the generated samples. Widely adopted metrics for gauging diversity and fidelity of image diffusion models include Inception Score (IS)~\cite{salimans2016improved}, Fr\'echet Inception Distance (FID)~\cite{heusel2017gans} and Kernel Inception Distance (KID)~\cite{bińkowski2021demystifying}. IS aims at capturing quality and diversity under a single metric by analyzing the distribution of labels obtained from a pre-trained classifier~\cite{szegedy2015rethinking}. FID computes the Fr\'echet distance between inception features~\cite{szegedy2015rethinking} derived from a set of real and synthesised images under the assumption that the feature vectors follow a Gaussian distribution. KID aims to improve on FID by relaxing the Gaussian assumption, directly measuring the Maximum Mean Discrepancy (MMD) between the two feature sets. Despite their popularity, inception-based metrics face several fundamental limitations. For instance, such metrics are reliant on the Inception-v3 pre-trained model, which could inherit biases depending on how the model was trained.

With the introduction of large-scale T2I diffusion models, generalization capabilities must also be taken into consideration. Zero-Shot FID~\cite{saharia2022photorealistic} provides a solution to this, evaluating FID for images generated from a subset (30k in Saharia et al.~\cite{saharia2022photorealistic}) of unseen prompts taken from the validation set and comparing them with reference images from the full validation set.

Conventional metrics for evaluating image quality include PSNR, SSIM, and LPIPs. PSNR measures the peak signal-to-noise ratio, SSIM measures the structural similarity, and LPIPs captures the perceived similarity based on learned features between two images. Recent methods also provide mid and high-level metrics that compare images, such as DreamSim~\cite{fu2023dreamsim} and CLIP~\cite{radford2021learning}. However, such metrics are only applicable when ground truth images are available for comparison, i.e., reconstruction tasks. In the context of evaluating generative models, these metrics are rarely used directly.

\paragraph*{Video Quality and Diversity.}

Naturally, the aforementioned image metrics have been extended to video, most notably using the Fr\'echet video distance (FVD) \cite{unterthiner2018towards} (e.g., implemented by the I3D network~\cite{carreira2017quo}). FVD is often reported in conjunction with (video) IS.
T2V datasets that are smaller than the large training sets, such as UCF101~\cite{soomro2012dataset} and
MSR-VTT~\cite{xu2016msr}, are often used to evaluate IS and FVD scores.

As noted in several works, e.g.,~\cite{brooks2022generating,blattmann2023align}, FVD is sensitive to the realism of individual frames
and motion over short segments, but it does not capture long-term realism. Unrealistic repetitions over time, for example, are not penalized. Moreover, as noted by~\cite{skorokhodov2022stylegan}, FVD is highly sensitive to small implementation differences, implying that reported results between papers are not always directly comparable. To address these challenges, many T2V approaches use human evaluation in addition to IS and FVD, as discussed below.

\paragraph*{Evaluating 3D Object Fidelity.}
The first option is to use FID and KID on the multiple renderings of the 3D models.
Given a suitable encoder, FID can be applied to the latent space of the 3D models to directly measure the quality and diversity.
P-FID, for example, uses PointNet++~\cite{qi2017pointnet++} as the encoder to obtain the latent representation from pre-sampled points on the surface of the 3D models.
For conditioned generation, such as single-view 3D reconstruction, ground truth multi-view images are often available. In this case, PSNR, LPIPS and other conventional image quality metrics are applied for the evaluation of novel view synthesis, whereas shape reconstruction metrics such as Chamfer Distance, F-Score, and Intersection of Union (IoU) are used to measure geometry accuracy.
In case the text or image condition does not have ground truth, which is the more general case, prompt fidelity becomes the main metric.
It measures the alignment of the generation with the conditional input, e.g.,~text and image.
Similar to FID, one can fall back to measure image-prompt fidelity (see the \textbf{Prompt Fidelity} paragraph below) by using the renderings of the generated 3D models as a proxy.
Recently, with the development of joint shape-text-image embedding space, e.g.,~in~\cite{zhao2023michelangelo} and~\cite{xue2023ulip}, it is possible to directly assess the shape-prompt fidelity,e.g.,~Shape-Image Score (SI-S) and Shape-Text Score (ST-S), by comparing the embedding of shape and the conditional inputs on a pre-trained shape-text-image embedding space.

\paragraph*{Evaluating Animated and Articulated Objects.}
Lei et al.~\cite{Lei2023} introduced the Instantiation Distance (ID) for measuring the similarity between a pair of articulated objects. This metric considers the part-level geometry and the object motion patterns.
%
The physical realism of generated motion can be evaluated by physics-based metrics. For this purpose, EDGE \cite{tseng2023edge} proposed a metric to measure the dynamic realism of the motion. 
Diversity (Div), Multi-Modality (MM) and Beat Alignment Score (BAS)~\cite{li2021learn}
are widely used metrics for the evaluation of 3D human dance motions. 
The core idea of BAS is to assess how close the kinematic beats ({i.e.}, directional changes of the per-joint velocity vectors) coincide with the music beats. 
Note, however, that in many dancing styles, kinematic beats deliberately do not coincide with music beats. 
Hence, BAS should be used in combination with other metrics (FID, Div and MM). 

\paragraph*{Human Evaluation.}
The standard metrics discussed above, including FID and FVD, are unreliable at best and should be considered proxy metrics. For this reason, many papers describing new diffusion models also perform human evaluation by running user studies. Oftentimes, the users are asked to subjectively evaluate image or video ``quality'' and ``faithfulness'' (e.g., with respect to a prompt) or they are asked to directly compare and rank two or more images based on some metric. These types of user user studies are difficult to replicate and resource intense to conduct, as discussed in more detail in the next section on open challenges.


\paragraph*{Prompt Fidelity.}
As conditional generation using text prompts is one of the most popular applications of the diffusion model, it is important to evaluate the faithfulness of the generated content with respect to the text prompt.
To measure the faithfulness of a generated image with respect to the text prompt condition, the average cosine similarity between prompt and image CLIP~\cite{radford2021learning} embeddings is often computed. Similarly, CLIPSIM~\cite{wu2022nuwa} measures the average CLIP similarity between generated video frames and text with T2V models. Similar metrics have also been proposed to measure alignment of edited images and human instructions~\cite{brooks2023instructpix2pix}.

\paragraph*{Identity Preservation.}
To assess multi-view facial identity (ID) consistency for generated 3D faces, the mean ArcFace~\cite{deng2019arcface} cosine similarity score between pairs of views of
the same synthesized face rendered from random camera poses is the de-facto standard metric. 
When reference images are available for generated identities---as is the case when finetuning diffusion models on a few images of a specific subject---the average pairwise cosine similarity between CLIP embeddings of generated and real (reference) images of the same subject can be used to evaluate the ID consistency.
Note, however, that this approach is not constructed to distinguish between different subjects that could have highly similar text descriptions (e.g.,
two different yellow clocks). 
For this reason, \cite{Ruiz2022DreamBoothFT} proposed to also evaluate ID consistency using the average pairwise cosine similarity between ViTS/16 DINO embeddings of generated and real images. The advantage of this metric is that DINO \cite{caron2021emerging} is not trained to ignore differences between subjects of the same class by design.
\section{Open Challenges}
\label{sec:open_challenges}

Despite the immense progress made in generative models in recent years, there remain a large number of unsolved problems. In this section, we detail some prominent ones, although many more exist and are described in greater detail in the references cited throughout this report. 

\paragraph*{Evaluation Metrics.}
As discussed before, standard image and video quality metrics, such as FID and FVD, are not always well aligned with human judgement, and often make undesirable assumptions about the distributional similarity across datasets. Despite this, available alternatives are not much better: comparative metrics like PSNR and LPIPs require matching ground-truth pairs, and conducting a user study can be costly, time-consuming, and often not much more informative. One direction for obvious improvement in this domain is the creation of better metrics that reliably and automatically assess the quality and diversity of generated content. These metrics should be applicable to a variety of data types, such 2D images, video, 3D, and dynamic 3D scenes. Furthermore, they should be aligned with human preferences, to enable faster, automated progress on method development without regular human intervention.

\paragraph*{Training Data.} Captioned image data is available in abundance, but labeled training data for 3D, video, and 4D generation is scarce. This makes it difficult to train higher-dimensional generative models. Open problems remain in the collection of large-scale datasets, whether they be explicitly in higher dimensions ({e.g.}, a large set of 3D models), or lower-dimensional projections of this data that can be used for learning high-dimensional priors ({e.g.}, a large dataset of multi-view images). Both options require additional research in their corresponding training protocols---large-scale 3D datasets will undoubtedly have domain gaps with real-world scenes (since it will be difficult to match the diversity of real-world scenes), and large-scale multi-view datasets may not encode the same distribution of 3D-consistent scenes. The ideal training protocol and dataset may even be a combination of these options, or trained in stages, as is common with inflated models. 

Another way to frame this decision is as an option of quality versus quantity. Can diffusion models leverage weaker supervision from a larger amount of training samples more effectively than stronger supervision from fewer samples? 
For instance, large language models trained on text leverage trillions of training samples, while image generation models are trained with only billions of images, and video models with even fewer. 
Multi-view datasets and 3D datasets provide more supervision than monocular video captures, but available datasets are at least yet another order of magnitude smaller in size. One may consider opting for a smaller, more informative dataset and instead choose to tackle the problem of data \emph{efficiency}: the ability of a generative model to generalize in low-data regimes.

\paragraph*{Efficiency.}
Inference-time sampling speed continues to be a concern for diffusion-based generative models. Where other generative methods (e.g., GANs) only require a single forward pass of a neural network, diffusion models can require up to thousands~\cite{dhariwal2021diffusion, ho2020denoising} of network evaluations to produce a single generated result. The sequential nature of the forward and reverse diffusion processes poses as a fundamental bottleneck for efficiency. A straightforward method for increasing sampling speed is by designing newer, lightweight efficient architectures, such that fewer time is spent on each denoising step~\cite{li2023snapfusion}. Distillation is another category of technique that aims to achieve the same result. Salimans et al.~\cite{salimans2022progressive} and Meng et al.~\cite{meng2023distillation} distill a pre-trained diffusion model into a model that requires fewer sampling steps. Consistency Models~\cite{song2023consistency}, TRACT~\cite{berthelot2023tract}, and BOOT~\cite{gu2023boot} take this to the extreme, proposing single-step generation distilled from pre-trained diffusion samplers. The question of how to generate the highest-quality output in the most efficient manner, however, remains an unsolved problem.

Training efficiency also poses a concern. The majority of existing models are currently trained at scale by corporations with large computational resource pools. Exploring solutions for reducing the training compute requirements remains a very valuable open problem, as training specialized models for targeted applications is currently out of reach for most researchers. 

\paragraph*{Controllability.}
As discussed in Maneesh Agrawala's blog post~\cite{Agrawala:blogpost}, most diffusion models are unpredictable black boxes. Most models are conditioned only on text prompts, and therefore require extensive prompt engineering to generate a desired image. Furthermore, text input alone often does not offer sufficient control to specify a particular image's exact appearance. This poor interface typically results in a lengthy trial-and-error process. A conversational interface is a suitable alternative. Explored briefly in InstructPix2Pix~\cite{brooks2023instructpix2pix} and further developed in Dall-E3~\cite{openaiDALLE3}, conversations and instructions allow for relative and intuitive adjustments to a current image. This process of aligning generations with human intentions largely remains an open problem. 

Other existing forms of control include matched prompt editing~\cite{hertz2022prompt}, conditioning (Sec.~\ref{sec:fundamental_conditioning_guidance}), customization (Sec.~\ref{sec:fundamentals:customization}), and guidance~\cite{epstein2023diffusion}. These are all effective strategies to make diffusion models more controllable and predictable, better enabling a user to achieve a desired image, but are far from a polished solution. Designing better and more intuitive interfaces around diffusion models that provide generalized control and predictable outputs remains an important open challenge.  

\paragraph*{Physical Grounding.}
Controlling, constraining, or guiding diffusion models to adhere to the rules of physics is also a promising direction. Certain modalities, such as 3D geometry and motion are heavily constrained by the physics of the underlying scene they model. In practice, these physical properties and rules can be embedded in the training process or network design to encourage generations to be more plausible. GANs, for example, have greatly benefited from being constrained by the physics of projective geometry, non-rigid object deformation, and physically based lighting to enable the training of 3D-aware models from single-view 2D image datasets~\cite{HoloGAN2019,chan2021pi} to generate articulated 3D characters~\cite{wu2023magicpony,bergman2022generative,Yuan2023} or re-lightable digital humans~\cite{deng2023lumigan,wang2023sunstage}, respectively. Exploring similar uses of diffusion models, {e.g.}, by leveraging a model's emergent geometry or correspondences~\cite{tang2023emergent,luo2023diffusion}, is an encouraging open challenge. 

\paragraph*{Robustness and Reproducibility.}
Despite the meteoric advancements in the field, catalyzed by seminal works such as DALL-E 2~\cite{openaiDALLE} and Imagen~\cite{saharia2022photorealistic}, a disconcerting discrepancy persists between officially publicized results and those obtained through independent reproduction, even when employing the officially released code. This gap is often bridged through manual prompt manipulation, hyper-parameter optimization, and random seed manipulation—practices that have regrettably become tacit yet universally accepted methods for achieving favorable outcomes. This inflation of success is especially detrimental in optimization-based paradigms, including text-to-3D generation, and is exacerbated by the prevailing competitive ethos within the research community. Consequently, the imperative to develop robust and reproducible algorithms that fundamentally ameliorate these issues cannot be overstated.



\section{Social Implications and Ethical Concerns} 
\label{sec:social}

\paragraph*{Distribution of Harmful Content.} Generative AI tools automate the process of content creation with photorealistic quality. This ability could be used to generate fake photos or videos of real people (DeepFakes). DeepFakes pose a societal threat that could be used for harm, either intentional or unintentional. Anyone with unrestricted access to generative AI tools could, for example, create an image or video of a celebrity with the intention of tarnishing their reputation. 

A number of measures can be put in place to prevent the distribution of harmful content. First, the ability of a user to generate harmful content should be prevented as best as possible. Most image generation models today, for example, prevent the generation of content depicting violence, gore, harassment, drugs, adult content, and generally offensive topics. Second, forensic techniques to detect DeepFakes are being developed by the AI community (e.g.,~\cite{agarwal2019protecting,rossler2019faceforensics,fox2021}). As the quality of generative AI tools advances, these types of efforts are becoming increasingly important but also challenging. 

\paragraph*{Copyright, Legal Exposure, and Privacy Concerns.} Foundation models are trained on billions of images, including content that may have been scraped without the consent of the creator, that may have been legally protected otherwise, or that may contain personally identifiable or sensitive information. Indeed, copyright infringement lawsuits have already been filed by artists against some of the companies behind foundation models for visual computing. 

\paragraph*{Bias and Fairness.} Similar to most machine learning methods, diffusion models can inadvertently learn and perpetuate biases present in their training data. This is a significant concern in terms of fairness and ethical considerations; further research to develop means to mitigate such biases is needed.

\paragraph*{Environmental Concerns.} Training foundation models requires substantial computational resources. For example, the relatively small StableDiffusion model was reportedly trained on 2.3 Billion images using 256 Nvidia A100 GPUs on Amazon Web Services for a total of 150,000 GPU-hours~\cite{Mostaque:2022}. According to unverified sources, OpenAI's large GPT-4 model was trained on about 25,000 Nvidia A100 GPUs for 90--100 days, costing more than \$100 Million. The carbon footprint of training GPT-4 is estimated to be between 12,456 and 14,994 metric tons of carbon dioxide equivalent~\cite{Ludvigsen:2023}. For comparison, the average yearly carbon footprint across humans on Earth is 4 tons. 
These considerations lead to significant concerns about the environmental impact of training foundation models.

\paragraph*{Economic Impacts.} As generative AI models become more integrated into various sectors, there are concerns about their impact on jobs and economic structures. For example, the ability to generate high-quality digital doubles or, more generally, digital humans as well as 3D scenes is expected to have a significant impact on video production, the visual effects, among many other industries. Some jobs in this creative industry may be displaced, but at the same time new creative job profiles will arise and new ways to monetize creative work will be arise through Generative AI. Generative AI also automates many tasks that workers in other areas do today. Although this is the case for many technologies, generative AI raises concerns about worker displacements at an accelerated rate.

\paragraph*{Explainability, Trust, and
 Accountability.} Machine learning and generative AI models learn correlations within their training data, not causality. Therefore, it may be challenging to understand why a model gave the answer it did. Moreover, generative AI models can synthesize new data, which may not always be truthful, or the models could be involuntarily trained on data that contains factual errors. These issues call trustworthiness into question. Some outcomes may even have legal consequences, raising questions of accountability.  

Researchers working in the field of generative AI must be cognizant of these issues. Moreover, policy and lawmakers, industry, and the research community must engage in a constructive dialogue to set meaningful legal boundaries for the unprecedented capabilities and dangers of emerging generative AI. 


\section{Discussion and Conclusion}\label{sec:conclusion}

In this state-of-the-art report, we have reviewed the theory and practice of emerging diffusion models for visual computing. We have introduced the basic mathematical concepts, implementation details and design choices of popular diffusion models, and important strategies for finetuning, sampling, conditioning and inversion, among others. Moreover, we have given a comprehensive overview of the rapidly  growing literature in this space, categorized by the type of generated medium, and discussed available datasets, metrics, open challenges, and social implications. Yet, this is an exploding field with papers and commercial models being released on a weekly or even daily basis. Thus, we hope that this STAR provides an intuitive starting point for the interested reader---artist, practitioner, and researcher alike.  

Although most of the numerous papers discussed in this STAR have been published in the last (few) year(s) and all important aspects of diffusion models have seemingly been addressed, many open challenges remain. Perhaps one of the highest-level goals of the field of visual computing is to amplify the creative potential of novice and advanced users alike and empower them to intuitively convert their imagination into an image, video, or 3D scene. The generative AI tools discussed in this STAR are a big step forward, but the community still has much work ahead to achieving this goal.

\section*{Acknowledgements} 

We thank the Stanford Graphics lab, the Snap Research team, and Guandao Yang for fruitful discussions.
R.P. was supported by the Stanford Graduate Fellowship.
W.Y. was supported by the SNF Postdoc Mobility fund.
V.G. and C.T. were supported by the ERC Consolidator Grant \textit{4DReply} (770784) and by the Saarbr\"ucken Research Center for Visual Computing, Interaction and Artificial Intelligence (VIA Center, \url{via-center.science}). 
T.D. was supported by the Israeli Science Foundation (grant
2303/20). 
A.K. was supported by BAIR and BDD sponsors, as well as DARPA Fiddler. 
M.N. was supported by an ERC Starting Grant Scan2CAD (804724) and the German Research Foundation (DFG) Research Unit ``Learning and Simulation in Visual Computing''.
B.O. was supported by the Deutsche Forschungsgemeinschaft (DFG, German Research Foundation) project 421703927 and the bidt project KLIMA-MEMES.
G.W. was supported by a PECASE from the ARO, by Samsung, and by the Stanford Institute for Human-Centered AI (HAI).

\bibliographystyle{eg-alpha-doi}
\bibliography{references,references_direct_3D}

\end{document}